\documentclass{article}

\PassOptionsToPackage{numbers, compress}{natbib}

\usepackage[main, final]{neurips_2026}

\usepackage{amsmath}
\usepackage{amssymb}
\usepackage[utf8]{inputenc} 
\usepackage[T1]{fontenc} 
\usepackage{hyperref} 
\usepackage{url} 
\usepackage{graphicx} 
\usepackage{booktabs} 
\usepackage{amsfonts} 
\usepackage{adjustbox} 
\usepackage{nicefrac} 
\usepackage{microtype} 
\usepackage{xcolor} 
\usepackage{xspace}
\usepackage{algorithm}
\usepackage{algpseudocode}
\usepackage{multirow}
\usepackage[normalem]{ulem}
\useunder{\uline}{\ul}{}
\usepackage{caption}
\usepackage{graphicx}
\usepackage{enumitem}

\newcommand{\method}{BigCell\xspace}
\newcommand{\methodtile}{\method\textsubscript{tile}\xspace}
\newcommand{\methodlarge}{\method\textsubscript{large}\xspace}
\newcommand{\ttimes}{$\times$}

\newcommand{\blfootnote}[1]{%
  \begingroup
  \renewcommand{\thefootnote}{}
  \def\thefootnotemark{}
  \makeatletter
  \makeatother
  \footnotetext{#1}
  \endgroup
}

\title{BigCell: Generating Gigapixel Whole-Slide Images}

\newcommand{\sbu}{\textsuperscript{1}}
\newcommand{\anl}{\textsuperscript{2}}
\newcommand{\uoc}{\textsuperscript{3}}
\newcommand{\uou}{\textsuperscript{4}}
\newcommand{\unc}{\textsuperscript{5}}

\author{%
    Srikar Yellapragada\sbu \And
    Alexandros Graikos\sbu \And
    Zilinghan Li\anl \And
    Kostas Triaridis\sbu \AND
    Tarak Nath Nandi\anl\textsuperscript{,}\uoc \And
    Karen Bai\sbu \And
    Beatrice Knudsen\uou \And
    Tahsin Kurc\sbu \And
    Prateek Prasanna\sbu \AND
    Rajarsi R. Gupta\sbu \And
    Ravi K. Madduri\anl\textsuperscript{,}\uoc \And
    Joel Saltz\sbu\textsuperscript{,}\unc \And
    Dimitris Samaras\sbu
}

\date{%
    \parbox{\textwidth}{\normalfont \centering
        \sbu Stony Brook University \hspace{1em} \anl Argonne National Laboratory \hspace{1em} \uoc  University of Chicago \\
           \uou The University of Utah \hspace{1em} \unc The University of North Carolina at Chapel Hill
    }
}

\begin{document}
\maketitle


\begin{abstract}

Whole-slide images (WSIs) are gigapixel-scale digital scans of histopathology tissue and serve as the primary basis for cancer diagnosis. Existing histopathology generative models operate at the patch scale, producing fixed-size tiles of at most 1024$\times$1024 pixels which encompass only a small fraction of the WSI. In contrast, text reports for clinical diagnoses and slide-level diagnostic labels -- the supervision that exists at scale -- operate at the whole-slide level. We present BigCell, the first approach to generate coherent whole-slide histopathology images at native gigapixel resolution. BigCell trains a flow-matching transformer at 2048$\times$2048 pixels on over 120,000 publicly available WSIs spanning multiple organ types. Conditioning on embeddings from pathology foundation models and using an inference-time sliding-window pipeline, BigCell generates coherent images of up to 32,768$\times$32,768 pixels, outperforming prior tile generators on both tile-level and slide-level fidelity metrics. To make gigapixel synthesis practical, we introduce an adaptive scheduling strategy that prioritizes important slide regions, speeding up image generation by over 3\ttimes with minimal quality degradation. Furthermore, we propose the first text-to-image synthesis at 8k resolution by training a separate model that produces dense conditioning grids directly from pathology text reports. We demonstrate that augmenting with synthetic samples improves slide-level few-shot classification performance by up to $18 \%$ over baselines trained on real data.
\blfootnote{
    Corresponding author: \texttt{myellapragad@cs.stonybrook.edu}.
    Code and models provided at the \href{https://histodiffusion.github.io/docs/projects/bigcell/}{project page}.
}
\end{abstract}

\section{Introduction}

Diffusion and flow-matching have enabled high-fidelity natural image synthesis at megapixel resolution \cite{diffusion4k,sana, pixartsigma}. Approaches that push beyond the megapixel-scale rely on the self-similarity of natural scenes, synthesizing fine-grained material detail from coarser inputs \cite{ultrazoom}. In this paper, we address a fundamentally different challenge: the synthesis of histopathology whole-slide images (WSIs). WSIs are high-resolution digital scans of tissue sections that often exceed 50,000\ttimes50,000 pixels, and are the primary source of information for pathologists to diagnose diseases, including cancer. Crucially, they exhibit diverse structures at each magnification: from cellular morphology at the finer scales to glandular and tumor architecture at the coarser scales. Moreover, distant regions of a WSI are tightly coupled as staining properties are globally consistent, and cell morphology must be compatible with the surrounding tissue type (e.g. tumor cells should not appear in a region of normal tissue). In natural scenes, by contrast, such global constraints are fewer and usually weaker. Pathologists rely on spatial structures that span thousands of pixels to perform accurate diagnoses, making the full resolution of a WSI key to its diagnostic value.

\begin{figure}[t]
    \centering
    \includegraphics[width=.98\linewidth]{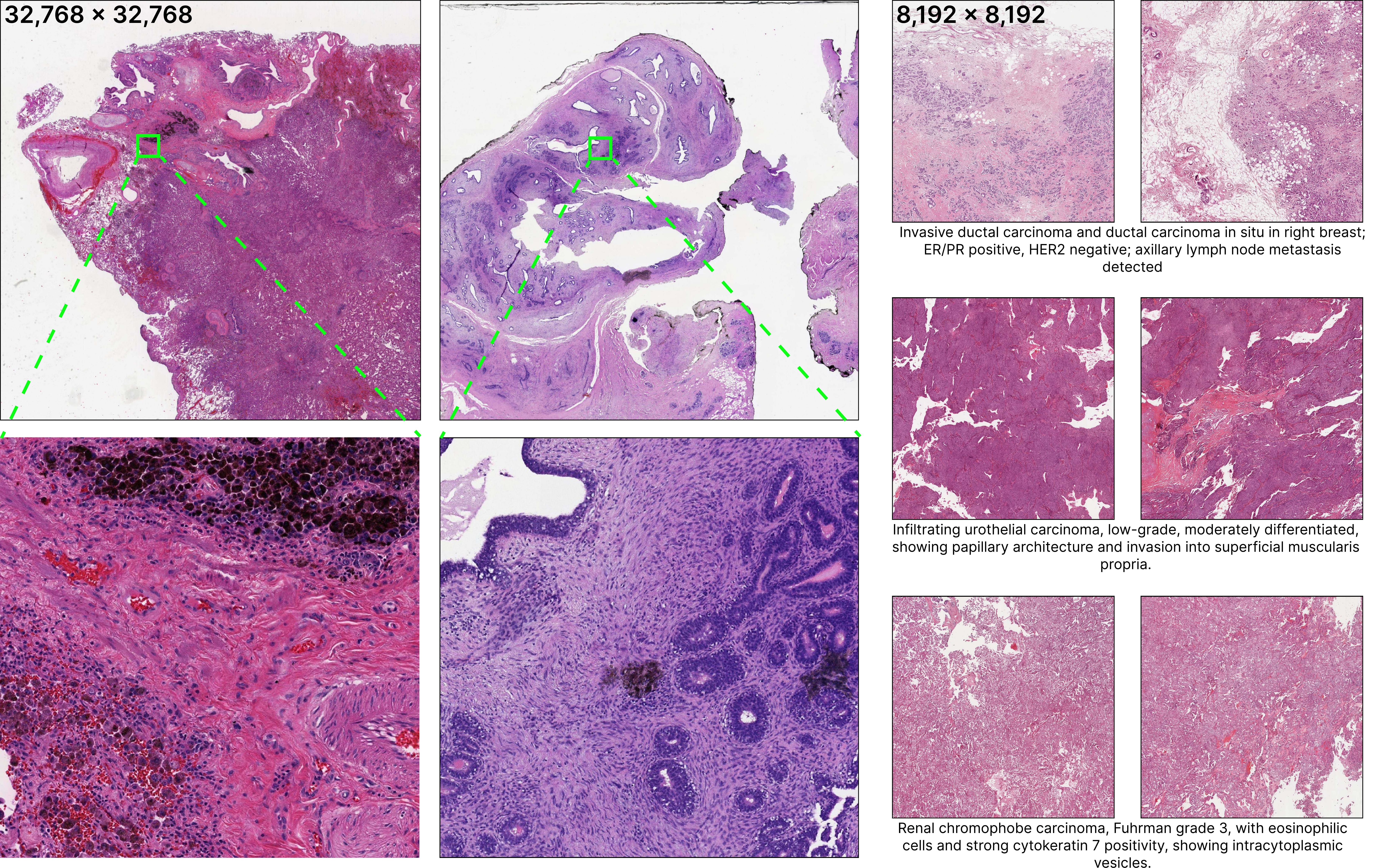}
    \caption{We present \textbf{synthetic} images generated using \method. Left: 32768\ttimes32768 images along with 2048\ttimes2048 crops, highlighting the realism achieved both at the macro-scale and cellular detail level. Right: Two  8192\ttimes8192 images per report generated from text. Best viewed zoomed in.}
    \label{fig:teaser}
\end{figure}

Existing generative methods in histopathology operate at the patch-level, synthesizing fixed-size tiles of WSIs (typically 256 to 1024 pixels) conditioned on semantic maps \cite{alfasly2025semantic, aversa2023diffinfinite}, textual descriptions \cite{bhosale2025pathdiff, unipath, mupad}, or learned representations \cite{cytosyn, lrdm,  pixcell, zoomldm}. One of the best performing models, PixCell \cite{pixcell}, generates tiles of up to 1024\ttimes1024 pixels natively, still orders of magnitude below whole-slide resolution. The scale gap is only one part of the problem: patch-level framing is often misaligned with pathology reporting. Pathologists work with WSIs, not tiles, and the clinical reports they produce describe findings at the level of the entire slide or case, with the latter comprising multiple slides. Pixel-level annotation at any meaningful scale is prohibitively expensive (\citet{lrdm} report 40 hours of pathologist time to annotate tiles from a single WSI with captions), while slide-level reports and labels are routinely produced in clinical practice. 

Modern computational pathology systems are increasingly being trained and evaluated at levels approaching whole-slides \cite{ticon, titan, moozy, prism, zimmermann2026mixed}. A collection of independently generated tiles cannot faithfully represent slide-level morphology, report-level diagnostic content, or long-range tissue organization. A generative model that aims to capture the distribution of pathology data and scale with the supervision that is widely available must therefore operate at the whole-slide level.

We present \method, the \emph{first} method to generate coherent whole-slide histopathology images at native gigapixel resolution in a single stage. Since training directly at this scale is computationally infeasible, we propose a scheme that progressively increases the base training resolution from 256 to 2048, controlling generation with embeddings from foundation models. Upon training a large-enough tile generator, we adopt a sliding-window inference-time algorithm \cite{lrdm} that synthesizes images beyond the training resolution, using the dense foundation model conditioning embeddings as a global blueprint, dictating exactly what tissue morphology belongs where.  Training on over 120,000 publicly available WSIs spanning multiple organ types, our approach scales to produce images of up to 32,768\ttimes32,768 pixels, the typical size of a small to medium WSI (Figure~\ref{fig:teaser}-left).

Sliding-window generation at gigapixel scale, however, is expensive: a single 32k pixel image requires denoising hundreds of overlapping windows at every timestep. We observe that not all windows require the same effort: complex regions rely heavily on the conditioning signal, while uniform regions (such as background or homogeneous stroma) only require coarse conditioning information to be synthesized accurately.
We measure this per-window importance as the difference between conditional and unconditional model predictions (akin to classifier-free guidance \cite{cfg}) and use it to reduce the number of generation steps in the identified low-importance regions. This strategy speeds up generation by over 3 times with minimal loss in image quality, is training-free, and requires only a single interpretable hyperparameter controlling the fraction of important windows.

To rigorously evaluate generation at this scale, we decouple local and global metrics: we measure local fidelity using standard patch-level similarity and Fréchet distance scores, and global coherence using slide-level foundation model similarity. When using previous histopathology tile generators for 8192\ttimes8192 image generation, we find that although they achieve high patch-level quality, their limited training resolution degrades the structural coherence across broader regions. Consequently, \method achieves superior macro-scale fidelity, reflected in higher foundation model similarity and improved bottom-5th-percentile patch-level similarity.

The above pipeline requires a real WSI to extract the conditioning embedding grid from. The most commonly available (and often the only) information about WSIs is the pathology report that accompanies it. Pathology reports describe tissue type, morphology, diagnosis, grade, and ancillary findings and are abundant in clinical archives. We train a flow-matching model, the conditioning generation model (CGM), that operates in the foundation model's representation space, generating dense conditioning grids suitable for \method's tile generator, directly from report text. The full grid is generated jointly, capturing the long-range dependencies present at this scale. The resulting pipeline produces 8192\ttimes8192 pixel histopathology images from free-form text descriptions, generating tissue whose alignment with the input report, as measured by both board-certified pathologists and multiple pathology foundation models, approaches that of real paired data (Figure~\ref{fig:teaser}-right). To our knowledge, \method is the first attempt to achieve text-controlled generation at this scale. 

The application we evaluate our generations on is synthetic data augmentation for computational pathology. We use \method to generate diverse and class-consistent variations of real training samples in various few-shot classification settings. On the TCGA-UT-8K \cite{titan} and PANDA \cite{panda} datasets, augmenting the real training set with synthetic samples from \method yields up to $18\%$ relative improvement in performance over baselines trained on real data alone. Beyond data augmentation, we expect the report-to-image technology to impact clinical practice in many ways. For example, it can empower clinicians to explore counterfactual scenarios by generating side-by-side comparisons of subtle diagnostic alternatives, which will be useful for difficult differential diagnoses; 
it can enhance diagnostic quality control by visualizing report findings to cross-check against the patient's slides. It can also improve histopathology education by generating teaching cases of rare morphologic variants. 


In summary, our contributions are: 
\begin{itemize}[noitemsep]
    \item[\textbf{(1)}] We present \method, the first method to generate whole-slide images at native gigapixel resolution, outperforming prior generative methods in global coherence and image fidelity. \\
    \item[\textbf{(2)}] We develop an inference-time strategy that uses classifier-free guidance magnitude to accelerate sliding-window generation, achieving over 3 times speedup with minimal quality degradation. \\
    \item[\textbf{(3)}] We show how learning the distribution of conditioning embeddings from clinical pathology reports enables the first text-to-pathology image generation at 8192\ttimes8192 resolution. \\
    \item[\textbf{(4)}] We demonstrate that synthetic samples from \method improve few-shot classification performance on two downstream datasets and convey the same diagnostic content as real tissue when verified by board-certified pathologists.
\end{itemize}

The entire \method pipeline is trained exclusively on publicly available datasets with no proprietary data. Our trained models and code are publicly released to facilitate further research.

\begin{figure}[t]
    \centering
    \includegraphics[width=\linewidth]{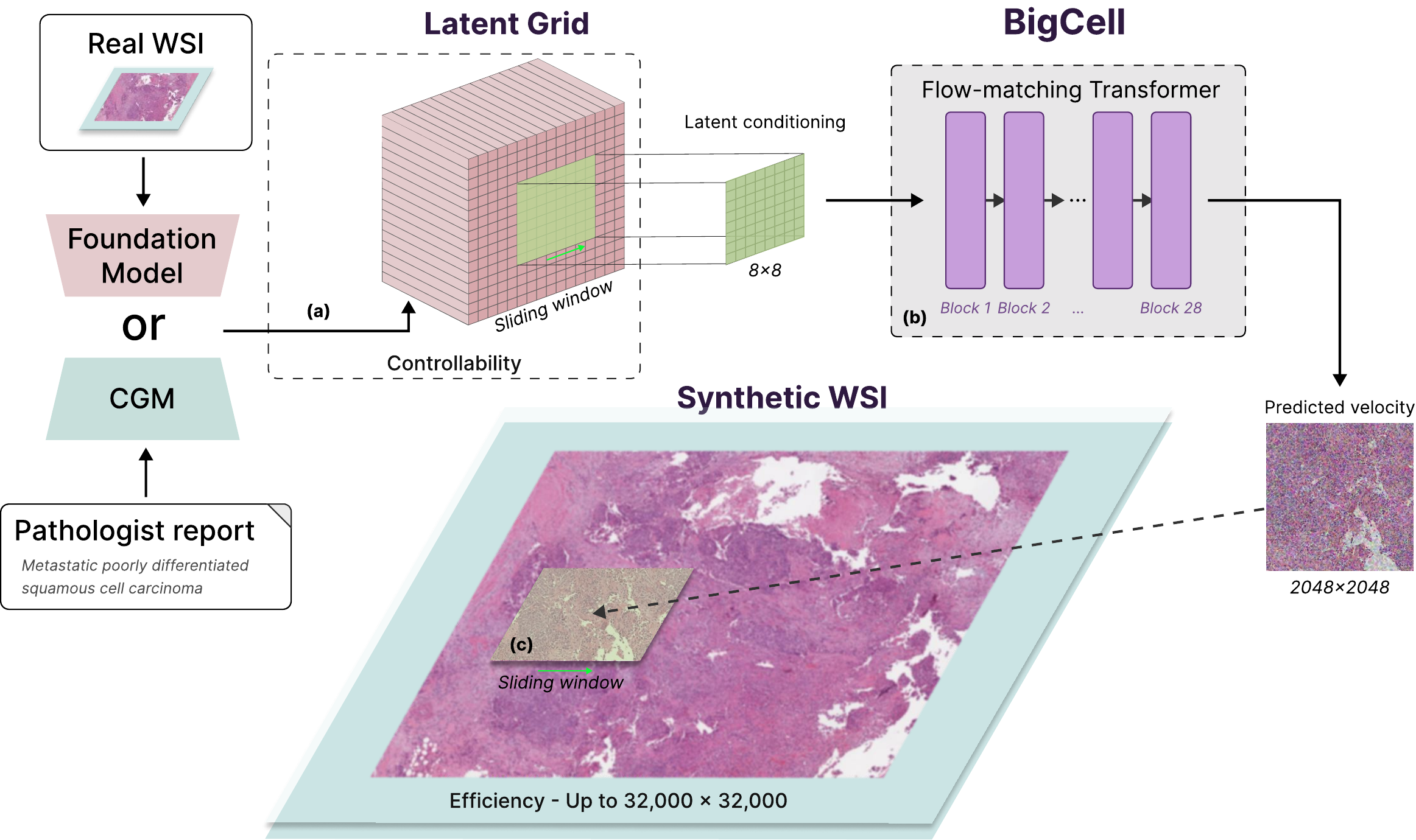}
    \caption{Overview of the \method pipeline. (a) We obtain a dense grid of conditioning embeddings either from a reference WSI or sampled using a model conditioned on text (CGM). This serves as the controllable blueprint that defines the large-scale structures of the whole-slide image. (b) The tile generator uses the local conditioning to predict how to transform the slide locally. (c) Using the tile generator on overlapping windows, we aggregate the per-window velocities into a global velocity field and iteratively denoise the noisy slide into a gigapixel image, allowing us to efficiently synthesize WSIs without explicitly training on their scale.}
    \label{fig:method}
\end{figure}

\section{Related Work}

\textbf{Natural image synthesis} has become synonymous with latent diffusion \cite{ldm} and transformer-based denoisers \cite{dit, sit}, with flow matching \cite{lipman2022flow, albergo2025stochastic} emerging as the prevailing approach. By scaling the latent diffusion recipe, prior work has pushed native generation to megapixel resolution (e.g., Stable Diffusion 3 \cite{sd3} and FLUX \cite{labs2025flux}). Pushing beyond the training resolution is typically handled by inference-time approaches that synthesize an image in overlapping windows \cite{multidiffusion,lee2023syncdiffusion}
or by fusing predictions from multiple resolution branches to reach higher resolutions \cite{freescale}. These approaches rely on the assumption that local appearance dominates fidelity, so distinct regions can be rendered independently without impacting global semantics. Whole slide images break this assumption: cellular morphology must match with the surrounding tissue type across thousands of pixels, and staining must remain globally consistent.

\textbf{Computational pathology} typically follows a two-stage approach for downstream tasks: a patch-level foundation model \cite{uni, virchow2, hoptimus1, conch, gensis} provides dense tissue embeddings, and a multiple instance learning (MIL) model \cite{abmil, clam} aggregates them into a slide-level prediction. More recently, the field is moving toward slide-level foundation models \cite{titan, prism, gigapath, moozy} and context-aware patch encoders \cite{ticon, zimmermann2026mixed} whose representations depend on their neighbors. These works have demonstrated that modeling dependencies beyond a single patch yields stronger representations for downstream tasks.

\textbf{Generative models in histopathology} have thus far operated at the patch scale, conditioned on semantic segmentation maps \cite{aversa2023diffinfinite, alfasly2025semantic}, text captions \cite{pathldm, bhosale2025pathdiff, unipath, mupad}, and embeddings from discriminative pathology foundation models \cite{lrdm, zoomldm, pixcell, cytosyn}. At the whole-slide level, \citet{harb2024diffusion} produce outputs of up to 65,536\ttimes65,536 pixels with a multi-stage cascade: an unconditional model samples a 512\ttimes512 thumbnail, which successive super-resolution stages progressively upsample. However, their WSI is sampled from a learned distribution only once, at the thumbnail level, and every subsequent stage refines a local, blurred crop of the stage below. 
DiffInfinite \cite{aversa2023diffinfinite} synthesizes large mask-image pairs by utilizing a per-pixel tissue-class segmentation mask with parallel patch diffusion. Distant patches are coupled only through the mask, which does not encode global properties such as staining, and the method has not been demonstrated at gigapixel scale. 

The representation-conditioned line of work is closest to our setting: LRDM \cite{lrdm} introduced sliding-window inference over an embedding-conditioned latent diffusion model, ZoomLDM \cite{zoomldm} added magnification-aware training and cross-scale consistency \cite{graikos2025fast}, and PixCell \cite{pixcell} progressively trained from 256 up to 1024 native pixels on over 69,000 slides, reaching 4096 through inference-time stitching. 
\method denoises the full latent canvas at 20\ttimes{} in a single stage, conditioning every window on a dense grid of 1536-dimensional foundation model embeddings that is either extracted from a reference slide or sampled jointly for the entire region by the CGM (Section~\ref{sec:cdm}).

\section{Method}


\subsection{Tile generator}
\label{sec:tile_generator} The core of \method is a flow-matching transformer (Figure~\ref{fig:method}(b)) that generates 2048\ttimes2048 pixel tiles conditioned on a dense grid of pathology foundation model embeddings. Following the latent diffusion framework, we operate in the latent space of a pretrained VAE \cite{flux-2}, which encodes a 2048\ttimes2048 tile into a 256\ttimes256\ttimes32 latent $z$. The generative model is a Scalable Interpolant Transformer (SiT) \cite{sit} parameterized as a velocity predictor $v_{\theta}(z_{t}, t, y)$, where $z_{t}$ is the noisy latent at timestep $t \in [0 ,1]$ and $y$ is the conditioning grid. We refer to this tile generator as \methodtile.

To condition the model, we follow the LRDM formula \cite{lrdm}, which uses features extracted from discriminative foundation models (Figure~\ref{fig:method}(a)). For each 2048\ttimes2048 tile, we extract UNI-2 \cite{uni} features from the 64 non-overlapping 256\ttimes256 patches, yielding an 8\ttimes8\ttimes1536 conditioning grid $y$, that is injected into the transformer via cross-attention layers.

\subsection{Sliding-window generation}
\label{sec:sliding_window}

Training directly at whole-slide resolution is infeasible. To generate images larger than the native 2048 resolution, we extend the LRDM framework \cite{lrdm} to flow matching (Appendix~\ref{appendix:averaging_velocitites}). Given a target image of $H \times W$ pixels, we first assemble a global conditioning grid $y \in \mathbb{R}^{(H/256) \times (W/256) \times 1536}$ of UNI-2 features, extracted from a reference WSI, or generated from text as in Section~\ref{sec:cdm}. At each denoising step, we run \methodtile independently on overlapping windows and aggregate the per-window velocities into a global velocity field, which we then use to take a deterministic Euler step (Figure~\ref{fig:method}(c)). After the final step, the VAE decodes the latent canvas to pixels.

The sliding-window procedure is training-free and can be applied to any tile generator. The fidelity of the final image, however, depends directly on the underlying tile generator's ability to faithfully render the conditioning blueprint. Using a higher native resolution, which captures a sufficiently large spatial context per velocity window, yields better fidelity at the gigapixel scale; in Appendix~\ref{ablation:tile_resolution}, we ablate how gigapixel image quality improves monotonically with native resolution from 256 up to 2048, with diminishing returns after 1024. We also demonstrate that \methodtile outperforms tile generators from prior work \cite{pixcell,cytosyn, mupad} across tile-level and slide-level metrics (Section~\ref{sec:results_tile_gen}). We refer to the sliding-window pipeline as \methodlarge.

\subsection{Efficient generation}
\label{sec:efficient_generation}

\methodlarge applies the tile generator on every window, which at gigapixel resolution, amounts to hundreds of forward passes per generation step, making it prohibitively expensive. A straightforward way to reduce this cost is to denoise only a subset of windows at each step, but this raises the question of which windows to prioritize.

We observe that windows have varying dependence on the conditioning signal. Some regions contain complex morphological structures (tumor boundaries, heterogeneous stroma) with the conditioning enforcing a particular cellular arrangement. Other regions (background, homogeneous stroma) are largely shaped by the unconditional prior, making the conditional and unconditional velocities nearly identical. We posit that such "simpler" windows also require fewer steps to be synthesized faithfully. Therefore, we propose to reuse the same velocity from the previous step as an approximation to the current velocity for less important windows, reducing the total number of forward passes. 

\noindent
\textbf{Importance via classifier-free guidance.} 
To distinguish between ``harder'' and ``easier'' windows, we measure how much each window relies on the conditioning during inference. During generation we already use classifier-free guidance (CFG) \cite{cfg}, which requires both the conditional $v_{\theta}(z_{ij}, t, y_{ij})$ and unconditional velocities $v_{\theta}(z_{ij}, t, \emptyset)$. The magnitude of their difference is therefore available for free, and we can use it to define a per-window importance score, $I_{ij}(t) = || v_{\theta}(z_{ij}, t, y_{ij}) - v_{\theta}(z_{ij}, t, \emptyset)||_{1}$. A high $I_{ij}(t)$ indicates that the conditioning signal strongly influences the velocity prediction. We provide a visualization of this importance signal in Appendix Figure \ref{fig:cfg_magnitude}.

The idea of allocating compute across tokens was proposed by \citet{wu2025importance} in the context of a single image. Their method discards information from low-importance tokens within a forward pass and cannot be used to share information across tiles. Our formulation achieves that by caching per-window velocities across denoising steps and reusing them when a window is deemed low-importance.

\noindent
\textbf{Adaptive scheduling.} Let $r \in (0,1]$ denote the fraction of windows to denoise per step. At the first denoising step, we compute $I_{ij}(t)$ for all windows. At each subsequent step, we compute velocities only for the top-$r$ most important windows from the previous step, and update their importance scores accordingly. Skipped windows reuse the cached velocity from the most recent step in which they were denoised. This strategy is training-free, requires only a single interpretable hyperparameter $r$, and reduces compute by approximately $1/r$. We ablate this tradeoff in Section \ref{sec:result_efficient_generation}, and show that we can achieve over 3 times speedup with minimal quality degradation.

\subsection{Conditioning generation model (CGM)}
\label{sec:cdm}

\methodlarge and \methodtile require a dense embedding grid $y$ as conditioning. In the default setting, $y$ is extracted from a reference WSI, which ties generation to existing slides. To enable novel generation from free-form text, we train a separate generative model that produces $y$ directly from a pathology report. We call this model the conditioning generation model (CGM).

\noindent
\textbf{Architecture and training.} The CGM is trained on a subset of WSIs paired with pathology reports describing tissue type, morphological findings, and diagnosis. Each report is encoded using the TITAN text encoder \cite{titan}, yielding a sequence of text embeddings $c$. CGM is also a flow-matching transformer that learns to sample from the conditional distribution $p(y|c)$, operating directly in the UNI representation space. Given a target UNI grid $y\in \mathbb{R}^{32 \times 32 \times 1536}$ that corresponds to a 8192\ttimes8192 pixel region, we apply the same linear interpolant formulation as \methodtile to train a model that samples $y$ from the given text tokens $c$.

\noindent
\textbf{Text-to-image generation.} At inference, we sample a UNI grid from the CGM conditioned on TITAN text embeddings, pass it to \methodlarge and produce an 8192\ttimes8192 pixel image. In Section~\ref{sec:result_text} we showcase how this two-stage process synthesizes gigapixel images from text that exhibit the desired properties specified in the given prompt.
To the best of our knowledge, our method is the first to achieve histopathology text-to-image generation at this scale.

\subsection{Synthetic data augmentation}
\label{sec:cdm_augmentation}

The CGM also enables a targeted form of data augmentation for downstream classification tasks. Given a real image from the training set, we can use the CGM to synthesize class-consistent variations by perturbing and denoising the embedding grid under a class-specific text prompt.

For a classification task with label set $\mathcal{L}$, we first construct $m$ synthetic pathology reports per class by prompting an LLM with the class name, yielding text embeddings $\{c_{l}\}_{l \in \mathcal{L}}$. Given a real training example $(x, l)$ with UNI grid $y_{0}$, we add noise to the embeddings up to an intermediate timestep $t'$,
and then denoise it back to $t=0$ using the CGM conditioned on $c_{l}$. The resulting UNI grid $\hat{y}_0$ is passed through \methodlarge to produce a synthetic image $\hat{x}$, and the pair $(\hat{x}, l)$ is added to the training set. This preserves the global structure of the original image while injecting class-consistent variations driven by the text prompt. For all augmentation experiments we use $t' = 0.5$.

\section{Experimental Setup}
\label{sec:experimental_setup}
\subsection{Data}
\label{sec:setup_data}

\noindent
\textbf{Training data.} Our training data consists of 120,703 WSIs from six publicly available sources: TCGA \cite{weinstein2013cancer}, CPTAC \cite{edwards2015cptac}, GTEx \cite{lonsdale2013genotype}, HistAI \cite{histai}, PTRC-HGSOC \cite{chowdhury2023proteogenomic}, and NLST \cite{national2011reduced}. We extract 2048\ttimes2048 pixel tiles at 20\ttimes magnification, and split the dataset at the slide level to obtain 15,996,332 tiles for training and 852,435 for evaluation.

\noindent
\textbf{CGM training data.} The CGM is trained on the subset of slides with paired pathology reports, drawn from TCGA diagnostic and HistAI, totaling approximately 30k patient-level reports. 
A single report may correspond to multiple slides from the same patient. We extract all 8192\ttimes8192 pixel regions across associated slides and construct target UNI grids, yielding 500k paired (grid, text embedding) samples for CGM training. We choose 8k as our operating point because at larger resolutions (e.g. 32k), the number of regions drops sharply, leaving too few samples to train a CGM. We hold out 150 patient-level reports and the corresponding slides as a test set for all text-conditioned evaluations.

\noindent
\textbf{Downstream classification data.} We perform k-shot classification on two datasets: PANDA \cite{panda}, a prostate cancer dataset with 6 ISUP grade classes, and TCGA-UT-8K \cite{titan}, which contains 8192\ttimes8192 pixel regions across the 32 target cancer types. 

\subsection{Metrics}
We evaluate generated images on: tile-level fidelity, slide-level coherence, report-level agreement, and text alignment:

\noindent
\textbf{Tile-level fidelity.} We extract non-overlapping 256\ttimes256 tiles from each generated slide and their corresponding reference slides. For each tile pair, we compute cosine similarity in the feature space of Virchow-2 \cite{virchow2} and H-Optimus-1 \cite{hoptimus1}. We report two statistics per slide: the average tile similarity ( V2\textsubscript{mean}, H1\textsubscript{mean} ), and the bottom 5th percentile tile similarity (V2\textsubscript{5th}, H1\textsubscript{5th}). The latter captures the prevalence of severe local artifacts that average similarity may miss. Additionally, we compute the Fréchet distances in both encoder spaces (V2\textsubscript{FD} and H1\textsubscript{FD}) on the tiles.

\noindent
\textbf{Slide-level fidelity.} To evaluate coherence beyond individual tiles, we compute cosine similarity between the generated and reference images in the feature space of slide-level foundation models: TITAN \cite{titan} and MOOZY \cite{moozy}.

\noindent
\textbf{Report-level agreement.} To complement embedding-based metrics with a measure that does not rely on pathology foundation models, we use MedGemma-4B \cite{medgemma}, a general-purpose medical VLM. We generate  structured reports, and measure the fraction of synthetic/real pairs whose reports agree on the organ, on the diagnosis, and on all three fields. (details in Appendix~\ref{appendix:medgemma}).

\noindent
\textbf{Text alignment.} For CGM-generated outputs, we evaluate alignment with the conditioning report by computing CLIP-style similarity between image and text embeddings under TITAN and PRISM \cite{prism}. We also report report-to-image retrieval recall@1 and recall@5 over the 150 report test-set.

\subsection{Baselines and evaluation procedure}

\noindent
\textbf{Tile generators.} We compare \methodtile against three prior histopathology tile generators: PixCell \cite{pixcell}, CytoSyn \cite{cytosyn}, and MuPaD \cite{mupad}. For each method, we use the publicly released model weights and apply the same sliding-window procedure described in Section \ref{sec:sliding_window} to generate 8192\ttimes8192 pixel images. The window size for each baseline matches its native resolution: 1024 for PixCell, 512 for MuPaD, and 224 for CytoSyn, with stride set to half the window size. We additionally compare against the large-image histopathology methods LRDM \cite{lrdm} and ZoomLDM \cite{zoomldm}, using their publicly released weights and their own large-image sampling procedures. We use 100 8192\ttimes8192 pixel ROIs sampled from the held-out test slides for evaluation and measure statistical significance via two-sided paired Wilcoxon signed-rank test vs best baseline (not applied to Fréchet distance).

\noindent
\textbf{Efficient generation.} We evaluate the adaptive window scheduling of Section \ref{sec:efficient_generation} at 32k resolution under matched compute budgets measured in number of function evaluations (NFEs). Evaluation is performed on 256 randomly sampled slides from the held-out test set, with one 32k pixel region generated per slide. The full baseline runs all windows at every step for 20 denoising steps. We compare against four operating points of our adaptive schedule with $r \in \{0.05, 0.09, 0.25, 0.41\}$. We compare against a \textbf{fewer-steps baseline}: the full sliding-window pipeline (all windows denoised at every step) but with the number of denoising (Euler) steps reduced to match the same NFE budget as our adaptive schedule.

\noindent
\textbf{Downstream classification.} For each dataset, we sample $k$ real training examples per class ($k \in \{1,2,4\}$), and generate synthetic variations using the CGM-based augmentation procedure described in Section \ref{sec:cdm_augmentation}. We compare the baseline trained on only the $k$ real images per class, and an augmented version trained on $k$ real + $k$ synthetic per class. For all variants, we use TITAN's vision encoder as the frozen feature extractor for both the real and synthetic samples. We then train a logistic regression classifier on top, and report the mean over ten random seeds of classifier training. Significance is assessed via a one-sided Wilcoxon signed-rank test across seeds on weighted kappa for PANDA and balanced accuracy for TCGA-UT-8K.

\begin{table}[ht]
    \centering
    \caption{Comparison at 8192\ttimes8192 resolution. V2: Virchow-2; H1: H-Optimus-1; FD: Fréchet distance (no test). $^{*}$: $p<0.001$, two-sided paired Wilcoxon vs.\ best baseline ($N=100$). Resolution- and data-matched comparisons are in Appendix \ref{ablation:matched}.}
    \vspace{1em}
    \begin{tabular}{lcccccccc}
    \toprule
    & \multicolumn{2}{c}{Slide-level Similarity} & \multicolumn{4}{c}{Patch-level Similarity} & \multicolumn{2}{c}{Fréchet Distance} \\
    \cmidrule(lr){2-3} \cmidrule(lr){4-7} \cmidrule(lr){8-9}
    Generator & TITAN\textuparrow & MOOZY\textuparrow & V2\textsubscript{mean}\textuparrow & V2\textsubscript{5th}\textuparrow & H1\textsubscript{mean}\textuparrow & H1\textsubscript{5th}\textuparrow & V2\textsubscript{FD}\textdownarrow & H1\textsubscript{FD}\textdownarrow \\
    \midrule
    PixCell                & {\ul .941}    & {\ul .900}    & {\ul .778}    & {\ul .632}    & {\ul .667}    & {\ul .509}    & 1023          & 696           \\
    CytoSyn                & .929          & .864          & .753          & .502          & .640          & .457          & 895           & 604           \\
    MuPaD                  & .646          & .755          & .462          & .258          & .342          & .196          & 981           & 641           \\
    \midrule
    LRDM                   & .531          & .693          & .513          & .334          & .395          & .247          & \textbf{731}  & \textbf{502}  \\
    ZoomLDM                & .708          & .801          & .646          & .464          & .477          & .325          & {\ul 815}     & {\ul 553}     \\
    \midrule
    \method & \textbf{.956}\textsuperscript{*} & \textbf{.973}\textsuperscript{*} & \textbf{.869}\textsuperscript{*} & \textbf{.737}\textsuperscript{*} & \textbf{.735}\textsuperscript{*} & \textbf{.586}\textsuperscript{*} & 907           & 621           \\
    \bottomrule
\end{tabular}

    \label{tab:8k_baselines}
\end{table}

\noindent
\textbf{Pathologist evaluation.} Three board-certified pathologists each completed 100 trials. In each trial, a query 8k image was shown alongside four candidate pathology reports; the pathologist selected the report believed to describe the query (multi-select with an \emph{unsure} option), and classified the query as real or synthetic. Of the 100 trials, 50 were real images and 50 were \method generations. The four candidate reports consisted of the true report and three randomly sampled reports from the test set.

\section{Results}

\subsection{Comparison against prior methods}
\label{sec:results_tile_gen}

In Table \ref{tab:8k_baselines}, we compare \methodtile against three prior tile generators -- PixCell-1024 \cite{pixcell}, CytoSyn-224 \cite{cytosyn}, and MuPaD-512 \cite{mupad} -- plugging each into the shared sliding-window pipeline to produce 8k images, as well as against the large-image histopathology methods LRDM \cite{lrdm} and ZoomLDM \cite{zoomldm}. \methodtile achieves the highest similarity to reference images under both slide-level encoders. Slide-level encoders aggregate information across the entire 8k image, and these gains indicate that \methodtile produces outputs whose global tissue arrangement more faithfully matches the conditioning grid. Note that prior tile generators were trained at different native resolutions; our ablation in Appendix \ref{ablation:tile_resolution} shows that higher native resolution leads to better fidelity at 8k, so these results reflect the combined effect of architecture and training resolution.

\methodtile also leads on patch-level cosine similarity under both Virchow-2 and H-optimus-1, in both mean and bottom-5th-percentile statistics. The gap is particularly large on the tail metrics: V2\textsubscript{5th} is 0.737 for \methodtile vs 0.632 for PixCell, 0.502 for CytoSyn, indicating substantially fewer regions of degraded local fidelity. LRDM and ZoomLDM attain the best Fréchet distances while obtaining the lowest agreement with the reference images; Fréchet distance is computed over unpaired 256\ttimes256 crops and does not measure agreement with the conditioning.

Table~\ref{tab:medgemma_8k} evaluates the same generations with report-level agreement using MedGemma-4B, which does not rely on pathology foundation model embeddings. \method again leads on all three fields, agreeing with the reference on the organ in 69\% of cases and on the diagnosis in 46\%, compared to 63\% and 42\% for PixCell. As all methods are evaluated on identical ROIs, these rates are meaningful relative to one another rather than in absolute terms.  Qualitative comparisons across the tile generators are shown in Appendix Figure \ref{fig:baselines_8k}.

\begin{table}[ht]
    \centering
    \caption{Report-level agreement at 8192\ttimes8192 resolution ($N=100$, same ROIs as Table~\ref{tab:8k_baselines}): fraction of generated/reference pairs whose MedGemma-4B reports agree on the tissue of origin, the diagnosis, and all three fields (organ, malignancy, diagnosis).}
    \vspace{1em}
    \begin{tabular}{lccc}
    \toprule
    Generator & Same organ\textuparrow & Same diagnosis\textuparrow & All three fields\textuparrow \\
    \midrule
    PixCell   & {\ul 63\%}     & {\ul 42\%}     & {\ul 38\%}     \\
    CytoSyn   & 54\%           & 26\%           & 25\%           \\
    MuPaD     & 41\%           & 17\%           & 17\%           \\
    \midrule
    ZoomLDM   & 61\%           & 40\%           & 36\%           \\
    LRDM      & 59\%           & 31\%           & 30\%           \\
    \midrule
    \method   & \textbf{69\%}  & \textbf{46\%}  & \textbf{41\%}  \\
    \bottomrule
\end{tabular}

    \label{tab:medgemma_8k}
\end{table}

\begin{figure}[ht]
    \centering
    \includegraphics[width=\linewidth]{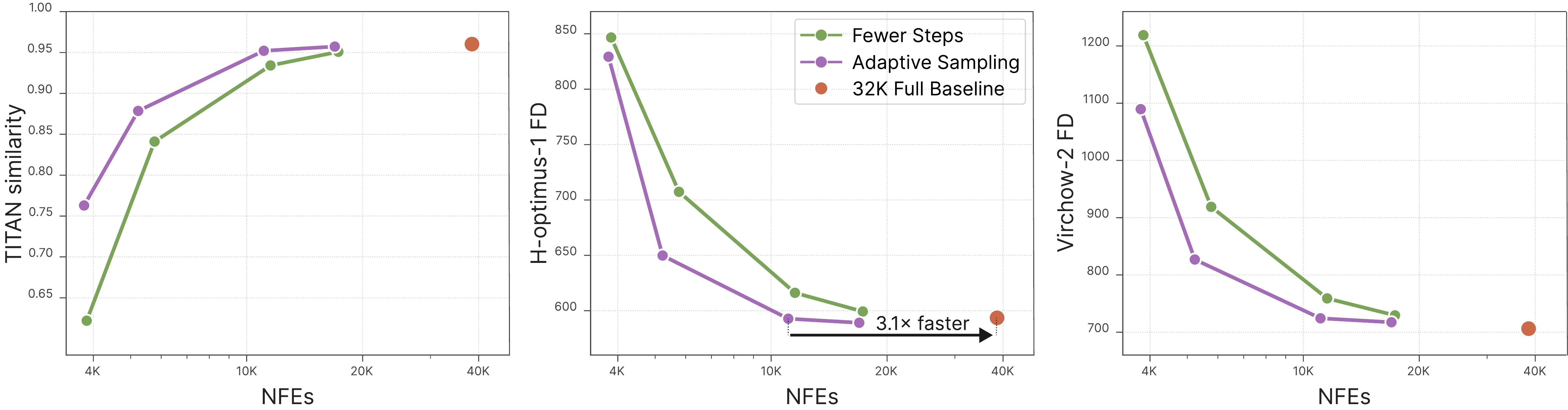}
    \caption{Adaptive window scheduling compared with simple, fewer-step sampling and the full 38440 NFE baseline at 32k resolution (N=256), evaluated at $r \in  \{0.05, 0.09, 0.25, 0.41\}$ . At $r=0.25$, we obtain a 3.1\ttimes speedup (15.6 h $\rightarrow$ 4.97 h on a single Intel Data Center GPU Max 1550).}
    \label{fig:32k_plots}
\end{figure}

\subsection{Efficient gigapixel generation}
\label{sec:result_efficient_generation}

We evaluate the adaptive window scheduling strategy of Section~\ref{sec:efficient_generation} at 32k resolution, and compare it against the fewer-steps baseline at matched compute budgets. Figure~\ref{fig:32k_plots} reports TITAN similarity and Fréchet distance with Virchow-2 and H-Optimus-1 as the encoders. Across all evaluated compute budgets, our adaptive schedule strictly improves over the fewer-steps baseline, with the gap being more pronounced at low compute. At NFE $\approx$ 4000, importance scheduling achieves a TITAN similarity of 0.76 while the fewer-steps baseline gets 0.63. At NFE $\approx$ 11000, our method yields a TITAN similarity of 0.951, matching the 0.957 similarity obtained at full compute (NFE = 38440). This leads to a 3.1 times speedup with negligible quality degradation, which, in wall-clock terms, reduces 32k generation from 15.6 hours to 4.97 hours on a single Intel Data Center GPU Max 1550.

\subsection{Text-conditioned generation}
\label{sec:result_text}

Table \ref{tab:text2img} shows the text-image alignment for 8k images generated by the CGM + \methodlarge for 150 held-out reports. We compare against the upper limit of real images paired with their corresponding reports. Generated images achieve text-image similarity that approaches the real paired data under both TITAN and PRISM encoders (TITAN: 0.487 vs 0.551; PRISM: 0.454 vs 0.499). As a sanity check, we also compute the similarity of randomly paired images and reports, which is significantly lower than what our model generated (TITAN: 0.118; PRISM: 0.203). This indicates that synthetic images faithfully reflect the content of the conditioning report. Retrieval recall further strengthens this conclusion: recall@5 under TITAN matches the real paired baseline (0.655), and recall@1 reaches 0.169 vs 0.211 for real data. 

\begin{table}[h]
    \centering
    \caption{TITAN and PRISM CLIP-style similarity and report-to-image retrieval recall on 150 held-out reports. CGM-generated images approach the alignment of real data under both encoders.}
    \vspace{1em}
    \begin{tabular}{cccccccc}
\toprule
 & \multicolumn{3}{c}{TITAN} & & \multicolumn{3}{c}{PRISM}            \\
 \cmidrule{2-4} \cmidrule{6-8}
Method & Sim & Recall@1 & Recall@5 & & Sim  & Recall@1 & Recall@5 \\
\midrule
Real data (paired)      & 0.551      & 0.211      & 0.655   &   & 0.499      & 0.148      & 0.430      \\
\method                 & 0.487      & 0.169      & 0.655   &   & 0.454      & 0.106      & 0.387      \\ 
\bottomrule
\end{tabular}
    \label{tab:text2img}
\end{table}

\begin{figure}[t]
    \centering
    \begin{minipage}[t]{0.38\linewidth}
        \vspace{0pt}
        \centering
        \captionof{figure}{Pathologist evaluation. Report-matching accuracy is comparable for real and synthetic queries.}
        \includegraphics[width=\linewidth]{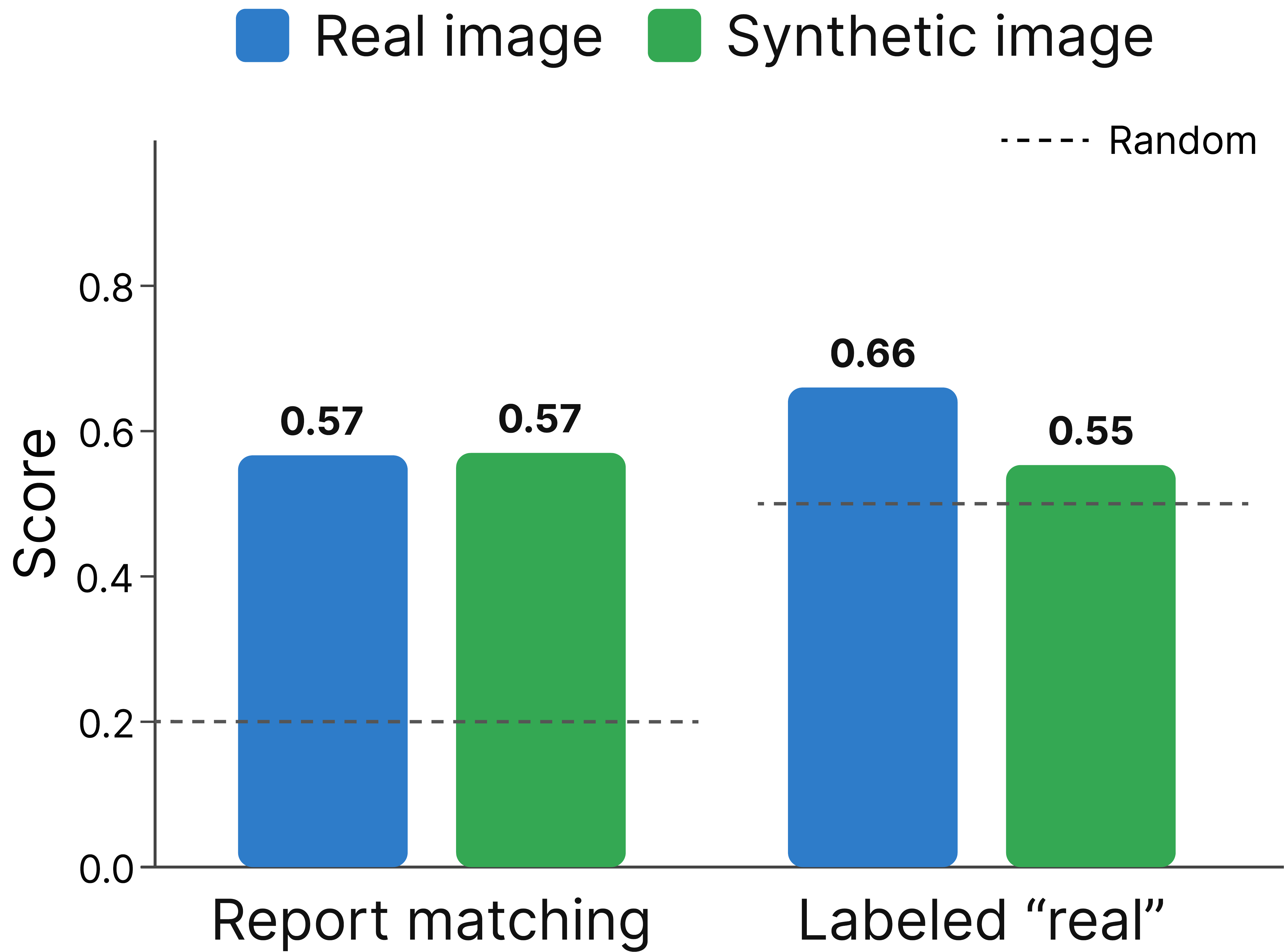}
        \label{fig:path_eval}
    \end{minipage}
    \hfill
    \begin{minipage}[t]{0.59\linewidth}
        \vspace{0pt}
        \centering
        \small
        \captionof{table}{Augmenting the real training set with synthetic samples from \method{} improves classifier performance over baselines trained on real data alone. * denotes $p < 0.05$.}
        \label{tab:augmentation}
        \begin{tabular}{ccccccc}
\toprule
&  & \multicolumn{2}{c}{TCGA-UT-8k} & & \multicolumn{2}{c}{PANDA} \\
\cmidrule{3-4} \cmidrule{6-7}
Shots & Split & Bal-acc & AUROC & & $\kappa$ & Bal-acc \\ 
\midrule
\multirow{2}{*}{K=1}   & Real        & .696                    & .970  & & .405                   & .304         \\
                       & +Syn & \textbf{.751\textsuperscript{*}} & \textbf{.979}  & & \textbf{.442\textsuperscript{*}}       & \textbf{.315}         \\
\midrule
\multirow{2}{*}{K=2}   & Real                   & .761            & .982     &     & .480       & .336         \\
                       & +Syn             & \textbf{.785\textsuperscript{*}}    &         \textbf{.984}   &      & \textbf{.531\textsuperscript{*}}       & \textbf{.358}         \\
\midrule
\multirow{2}{*}{K=4}   & Real                   & .801            & .987     &     & .493       & .375         \\
                       & +Syn & \textbf{.804}            & .987     &     & \textbf{.583\textsuperscript{*}}       & \textbf{.402}         \\
\bottomrule
\end{tabular}
    \end{minipage}
\end{figure}

\noindent
\textbf{Pathologist evaluation.}
In Figure \ref{fig:path_eval}, we present the results of the pathologist evaluation. Pathologists match synthetic queries to their source reports as accurately as real queries ($0.57$), indicating that synthetic images convey the same diagnostic content as real ones. The matching accuracy is upper-bounded by the task design: each report describes a patient but covers multiple distinct 8k regions, so a distractor with a similar diagnosis can fit the query as well as the true report. On the real-vs-synthetic task, pathologists label synthetic queries as real $55 \%$ of the time and real queries as real $66 \%$ of the time. Overall detection accuracy is 55.3\% (95\% CI 50–61\%), and a  permutation test that shuffles the real/synthetic labels across trials shows that this gap is not significant. Inter-rater agreement is at chance (Cohen's $\kappa = 0.020$), indicating that the three pathologists in our study did not reliably distinguish \method outputs from real tissue.

\noindent
\textbf{Synthetic data augmentation.}
Table \ref{tab:augmentation} shows few-shot classification performance on TCGA-UT-8K and PANDA datasets with and without CGM-based augmentation. Note that PANDA is held out from CGM training, providing an out-of-distribution evaluation. Augmentation improves performance across both datasets and all shot counts, with relative gains of up to $18 \%$ in PANDA and $7 \%$ in TCGA-UT-8K. Augmenting directly at the scale at which the classifier operates is key to improving prediction accuracy. With a tile-level model that independently augments tiles in a slide, we are not guaranteed to maintain the large-scale structures that may be critical in making the correct prediction.

\section{Limitations}
\label{sec:limitations}

We note three key limitations. First, text-conditioned generation is restricted to 8192\ttimes8192 pixel resolution (see Sec.~\ref{sec:setup_data}). Scaling text-to-image synthesis to gigapixel resolution would require substantially larger paired datasets than currently publicly available. Second, gigapixel generation remains expensive. Our adaptive scheduling reduces 32k generation from 15.6 hours to 4.97 hours on a single GPU, but the absolute runtime is still impractical for high-throughput applications. Third, synthetic histopathology data carries a risk of misuse; generated images are intended for research only and should not substitute for real tissue in clinical or diagnostic settings.

\section{Conclusion}

We presented \method, the first approach to generate coherent whole-slide images at native gigapixel resolution. \method produces images of up to 32,768\ttimes32,768 pixels, enabled by our adaptive scheduling strategy that speeds up generation by over 3\ttimes. Generated images outperform prior methods in both tile-level and slide-level fidelity metrics; we further guide it with text using the conditioning generation model that produces dense grids of pathology foundation model embeddings directly from reports. \method demonstrates the first text-to-image generation at 8192\ttimes8192 resolution, with text-generated samples improving few-shot classification performance on two downstream datasets and conveying the same diagnostic content as real tissue when verified by board-certified pathologists.

Our work opens promising directions for future research: generating a WSI from its report would allow imputation of slides for historical archives that predate digitization and for regions without scanning infrastructure. Beyond text, our method can support other modalities, such as spatial transcriptomics or multiplex imaging, enabling multi-modal generation. Providing the means to multi-modal generative models in digital pathology at the WSI-scale is paramount to building digital twin systems with the aim of accurately simulating medical scenarios and intervention outcomes.

\begin{ack}
This research was partially supported by National Science Foundation (NSF) grants IIS-2123920, IIS-2212046, National Institutes of Health (NIH) grant 1R21CA258493-01A1, as well as  Stony Brook Profund 2022 seed funding and private support from Bob Beals and Betsy Barton. This research used resources of the Argonne Leadership Computing Facility, a U.S. Department of Energy (DOE) Office of Science user facility at Argonne National Laboratory and is based on research supported by the U.S. DOE Office of Science-Advanced Scientific Computing Research Program, under Contract No. DE-AC02-06CH11357.
\end{ack} 

\bibliographystyle{abbrvnat}
\bibliography{ref}


\newpage
\appendix
\appendix

\begin{figure}
    \centering
    \includegraphics[width=\linewidth]{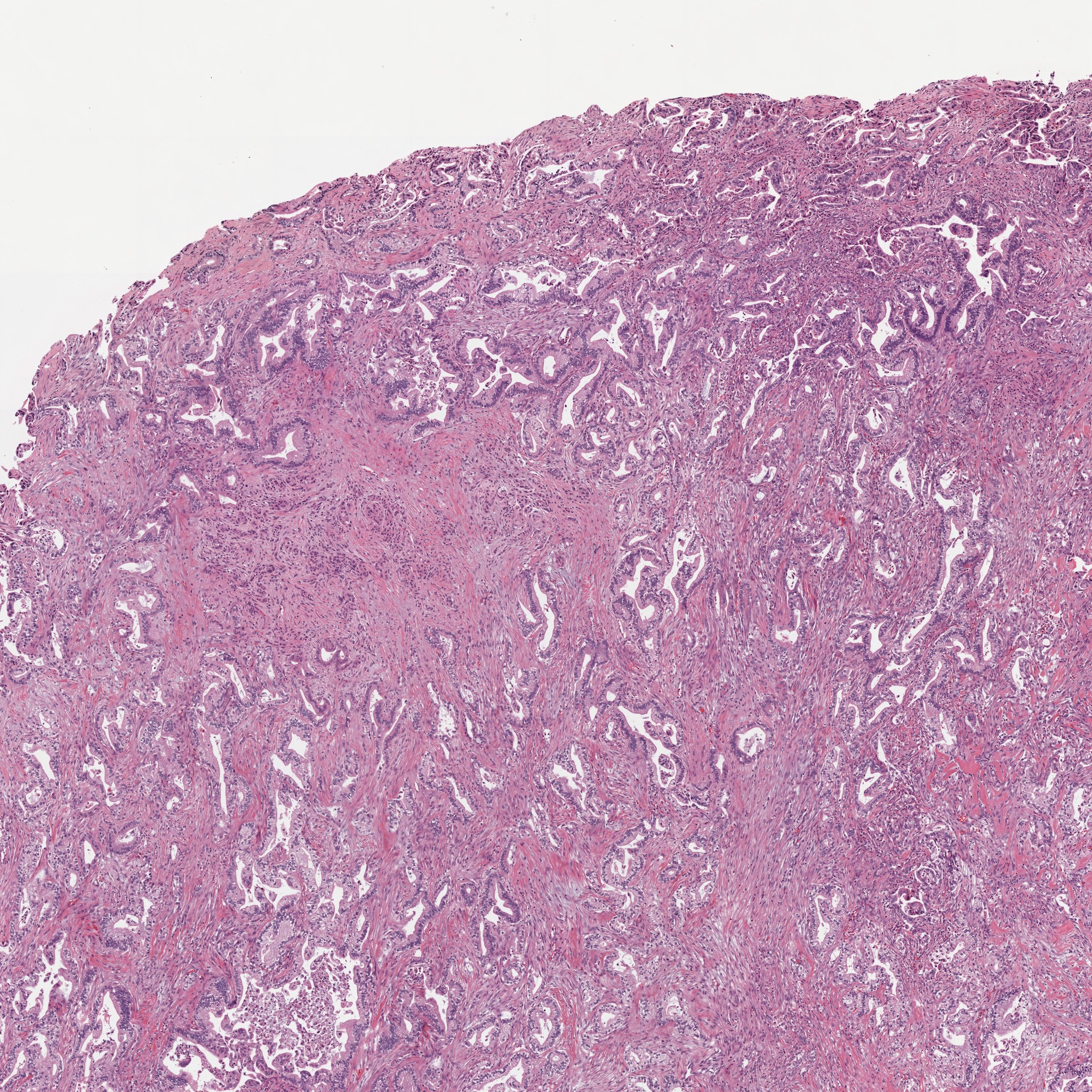}
    \caption{Synthetic 8192\ttimes8192 image generated by \method from a pathology text report, showing extensive pancreatic ductal adenocarcinoma.}
    \label{fig:8k_image_pancreas}
\end{figure}

\section{Motivation for whole-slide generation}

Generating a coherent WSI requires modeling biological dependencies that span the entire slide. Existing generative methods in histopathology operate at the patch level, and any approach that composes such outputs into a WSI must explicitly coordinate them across the canvas. Without such coordination, independent patch generation produces incoherent or biologically implausible outputs even when each patch is locally realistic.

\noindent\textbf{Diagnosis depends on the global spatial structure.} Pathologists interpret local findings in relation to the surrounding tissue. Distinguishing invasive carcinoma from in situ disease, for example, requires continuity across the tumor-stroma interface and assessment of growth patterns over large regions. Patches remove this context and cannot support reliable diagnostic reasoning.

\noindent\textbf{Pathology reports aggregate features at the slide level.} Key elements of a clinical report (tumor grade, margin status, extent of invasion, spatial heterogeneity) are defined by patterns spanning large portions of the slide. These features cannot be inferred from independently generated patches because they depend on consistent relationships across distant regions.

\noindent\textbf{Clinical trust requires global coherence.} For a generated image to be useful in a clinical workflow, staining, morphology, and tissue organization must remain consistent across the entire slide. Even subtle inconsistencies introduced by patch-based synthesis can undermine confidence and limit adoption.

\section{Clinical Applications}

Generating high-resolution synthetic 8192\ttimes8192 images of regions of interest (ROIs) based on complex text descriptions would be highly valuable for a wide variety of clinical and translational research applications. This technology can be utilized to generate controllable examples of "ground truth" for algorithmic development, iterative refinement, and benchmarking quality control for computational pathology. Figure \ref{fig:8k_image_pancreas} is a synthetic example that shows extensive pancreatic ductal adenocarcinoma (PDAC) surrounding a small focal area of chronic pancreatitis with residual pancreatic acinar tissue and parenchymal fibrosis. BigCell can be used to readily generate multiple examples of pancreaticobiliary cancers invading the duodenum, peripancreatic adipose tissue, and regional lymph nodes that vary in terms of extent of invasion (cancer staging), grade/differentiation (growth pattern and glandular formation), and inclusion of rare variant cell morphology. 

Considering the diversity of cancer as a disease and its uniqueness in each patient, this technology will be leveraged to visually examine common and rare morphologic findings side-by-side with synthetic hypothetical examples from massive real datasets to serve as a robust educational tool for residents, fellows, and other trainees. Having access to an inexhaustible library of histologic variants of cancer and other rare histological entities will be invaluable for the development of their professional microscopic examination and diagnostic skills, as well as support their preparation for board certification exams. Beyond the conventional PDAC shown in Figure \ref{fig:8k_image_pancreas}, \method can be utilized to generate synthetic examples of adenosquamous (glandular and squamous components), signet ring cell (poorly cohesive cancer cells with cytoplasmic mucin), and undifferentiated (lacking gland formation +/- osteoclast-like giant multinucleated cells) variants of PDAC, which would be extremely helpful for improving diagnostic recognition and slide navigation to salient ROIs for these kinds of rare entities associated with more aggressive disease than conventional PDAC. Similarly, \method can generate diverse examples to study cancer evolution and potentially predict scenarios associated with varying disease progression and clinical outcomes.

\section{Extending sliding window generation to flow-matching}
\label{appendix:averaging_velocitites}
Here we discuss the extension of the sliding window algorithm of \cite{multidiffusion,lrdm} to flow-matching models, where the trained model is predicting velocities instead of noise. 

At each generation step, we have a `noisy canvas' $x_t \in \mathbb{R}^{H \times W \times D}$, which is larger than what the tile-level model is trained with. Our goal is to compute the velocity for the entire canvas $v_{t}$, by using only the local velocities $v^i_t = v_{\theta}(x^i_t, t, c^i)$ for all windows $x^i_t$.

We denote the window extraction operation as $P_i x_t = x^i_t$, and we assume that the windowed global velocity should be similar to the local velocities, i.e., $P_iv_t = v^i_t$.

Following MultiDiffusion \cite{multidiffusion} and LRDM \cite{lrdm}, we pose the task of finding the global velocity as minimizing the least squares problem 
\begin{equation}
    v^{*}_{t} = \arg\min\limits_{v_t} \sum\limits_i\ \lVert W^i \odot (P^iv_t - v^i_t) \rVert_2^2,
    \label{eq:least_sq_velocity}
\end{equation}
where $W^i$ is a windowing function that handles overlapping pixels (for example uniform or Gaussian kernel). Looking at an individual location $p$, which indexes both the global and local velocities, we can rewrite the problem as
\begin{equation}
    v^{*}_{t}(p) = \arg\min\limits_{v_t} \sum\limits_i\  w^i(p)^2 (v_t(p) - v^i_t(p)) ^2,
    \label{eq:least_sq_velocity_pixel}
\end{equation}
which is straightforward to solve by setting 
\begin{equation}
    \frac{\partial v^*_t(p)}{\partial v_t(p)} 
    = 2\sum\limits_i w^i(p)^2 (v_t(p) - v^i(p))
    = 0.
    \label{eq:least_sq_zero}
\end{equation}
The solution
\begin{equation}
    v^*_t(p)
    = \frac{\sum\limits_i w^i(p)^2v^i(p)}{\sum\limits_i w^i(p)^2}
    \label{eq:least_sq_sol}
\end{equation}
tells us that for the uniform weighting $w^i(p) = 1$ that we use in our experiments, the global velocity can be approximated by averaging all (overlapping) local velocities at every location $p$ in the global canvas. This solution can be seen as the flow-matching extension of the result of \cite{multidiffusion}, which was applied to a diffusion model.

Having computed the global velocity, we can continue sampling by just replacing its value in our (Euler) sampler as 
\begin{equation}
    x_{t+\Delta t} = x_t + \Delta t\ v^*_t.
    \label{eq:global_euler}
\end{equation}

\section{Implementation details}
\label{appendix:implementation_details}

\subsection{Tile generator}

\methodtile is an SiT-XL/2 \cite{sit} flow-matching transformer with 830M parameters. We condition the model with UNI-2 \cite{uni} features via cross-attention in every block and operate in the Flux-2 VAE \cite{flux-2} latent space, which compresses 2048\ttimes2048\ttimes 3 images into 256\ttimes256\ttimes32. 

Table~\ref{tab:tensor_shapes} traces the tensor shapes through the pipeline. The generated object is the latent $z$; the final resolution is restored by a single decode through the Flux-2 VAE at a fixed 8\ttimes{} spatial factor. The conditioning grid $y$ enters only through cross-attention and is not part of the decode path. For larger canvases, the same relationship holds by scaling both tensors: each sliding window denoises a $256 \times 256 \times 32$ latent region conditioned on the corresponding $8 \times 8 \times 1536$ slice of the global grid, the per-window velocities are aggregated into a global velocity field (Section~\ref{sec:sliding_window}, Appendix~\ref{appendix:averaging_velocitites}), and the canvas is decoded once after the final denoising step.

\begin{table}[h]
    \centering
    \caption{Tensor shapes in \method for a single 2048\ttimes2048 tile and a 32,768\ttimes32,768 canvas. At 32k, each sliding window operates on a $256 \times 256 \times 32$ latent region with the corresponding $8 \times 8 \times 1536$ slice of the conditioning grid.}
    \vspace{1em}
    \begin{tabular}{llcc}
    \toprule
    Step & Tensor & 2048\ttimes2048 tile & 32,768\ttimes32,768 canvas \\
    \midrule
    Input / output image          & image $x$                      & $2048 \times 2048 \times 3$ & $32768 \times 32768 \times 3$ \\
    VAE encode / decode           & latent $z$                     & $256 \times 256 \times 32$  & $4096 \times 4096 \times 32$  \\
    UNI-2 on 256\ttimes256 patches & conditioning grid $y$          & $8 \times 8 \times 1536$    & $128 \times 128 \times 1536$  \\
    Denoising (per window)        & velocity $v_\theta(z_t, t, y)$ & $256 \times 256 \times 32$  & $256 \times 256 \times 32$    \\
    \bottomrule
\end{tabular}

    \label{tab:tensor_shapes}
\end{table}

We train under the linear interpolant formulation of SiT with velocity prediction. Following PixCell \cite{pixcell}, we progressively increase the training resolution from 256 to 2048 pixels. At each stage, we initialize the model from its previous stage weights. We also chose a shifted schedule for higher resolutions, as proposed by Stable Diffusion 3 \cite{sd3}. For training at lower resolutions (256, 512, 1024), we extract image crops and smaller UNI sub-grids from the same 2048\ttimes2048 dataset. We summarize our design choices and configuration in Tables~\ref{tab:bigcell_training} and \ref{tab:training_config_tile}.

\begin{table}[h]
    \centering
    \caption{Progressive training schedule for \methodtile. We train for one epoch at each resolution, and initialize the model from the previous stage weights.}
    \vspace{1em}
    \begin{tabular}{ccccc}
\toprule
Resolution & Batch size & Conditioning & lr   & t schedule          \\ 
\midrule
256        & 4800       & $1 \times 1536$          & 2e-4 & Uniform             \\
512        & 2400       & $4 \times 1536$          & 1e-4 & Uniform             \\
1024       & 720        & $16 \times 1536$         & 1e-4 & Shifted, $\alpha=3$ \\
2048       & 360        & $64 \times 1536$         & 1e-4 & Shifted, $\alpha=3$ \\
\bottomrule
\end{tabular}
    \label{tab:bigcell_training}
\end{table}

\subsection{CGM}

The CGM is a DiT-DH-Base \cite{repae} flow-matching transformer operating directly in UNI-2 CLS token space, with 362M parameters. The model uses RoPE positional embeddings, RMSNorm, and SwiGLU activations. We condition the model with $128 \times 768$ unpooled TITAN text embeddings via cross-attention. We train under the linear interpolant formulation with velocity prediction and logit-normal timestep sampling, with a time-distribution shift of $\sqrt{(32\times 32\times 1536)/4096}$ to account for latent dimensionality. The CGM training configuration is summarized in Table~\ref{tab:training_config_tile}.

\begin{table}[h]
    \centering
    \caption{Training configuration for \methodtile and CGM}
    \vspace{1em}
    \begin{tabular}{ccc}
\toprule
Component          & \methodtile & CGM                \\ 
\midrule
Optimizer          & AdamW                      & AdamW              \\
$\beta_1, \beta_2$ & 0.9, 0.999                 & 0.9, 0.95          \\
Weight decay       & 0                          & 0                  \\
EMA decay          & 0.9999                     & 0.9995             \\
Gradient clipping  & 1.0                        & 1.0                \\
Warmup             & 5,000 steps (linear)       & 40 epochs (linear) \\
CFG dropout        & 0.1                        & 0.1                \\
\bottomrule
\end{tabular}
    \label{tab:training_config_tile}
\end{table}

\subsection{Compute details}

We use an Intel GPU cluster for training. Each node contains 6 Intel Data Center GPU Max 1550 and 2 Intel Xeon CPU Max 9470C processors. We use PyTorch with the Intel Extension for PyTorch (IPEX) library, and perform distributed training using 12 ranks per node via MPI. In total, we consume 8,000 node-hours for this paper, including exploration and failed runs.

\subsection{Data pipeline}

We use the TRIDENT library \cite{trident} for tissue segmentation and patch extraction at 20\ttimes magnification. UNI-2 grids, VAE features and TITAN features were pre-computed and stored on disk. We did not apply stain normalization or data augmentation during \methodtile training.

\subsection{Sampling details}
For the tile generator, we use the Euler ODE sampler with 20 sampling steps and a classifier-free guidance scale of 2.0. We apply the same configuration for all baseline tile generators (PixCell, CytoSyn, MuPaD) for fair comparison. For the CGM, we sample UNI-2 grids from text reports using the Euler ODE sampler with 50 sampling steps and a classifier-free guidance scale of 2.0.

\section{Ablations}

\subsection{Effect of tile generator resolution}
\label{ablation:tile_resolution}

\begin{table}
    \centering
    \caption{Training cost across native resolutions, measured on the Intel GPU cluster. Cost scales quadratically, making training at 4096 impractical. The 4096 row is a single-iteration benchmark; we did not train a model at this resolution.}
    \vspace{1em}
    \begin{tabular}{cccc}
\toprule
Resolution & Batch size & Sec/Iter & Compute/Sample \\
\midrule
1024 & 720 & 1.35 & 1.0$\times$ \\
2048 & 360 & 5.59 & 8.3$\times$ \\
4096 & 360 & 65    & 96$\times$ \\
\bottomrule
\end{tabular}

    \label{tab:resolutions_time}
\end{table}

\begin{figure}
    \centering
    \includegraphics[width=\linewidth]{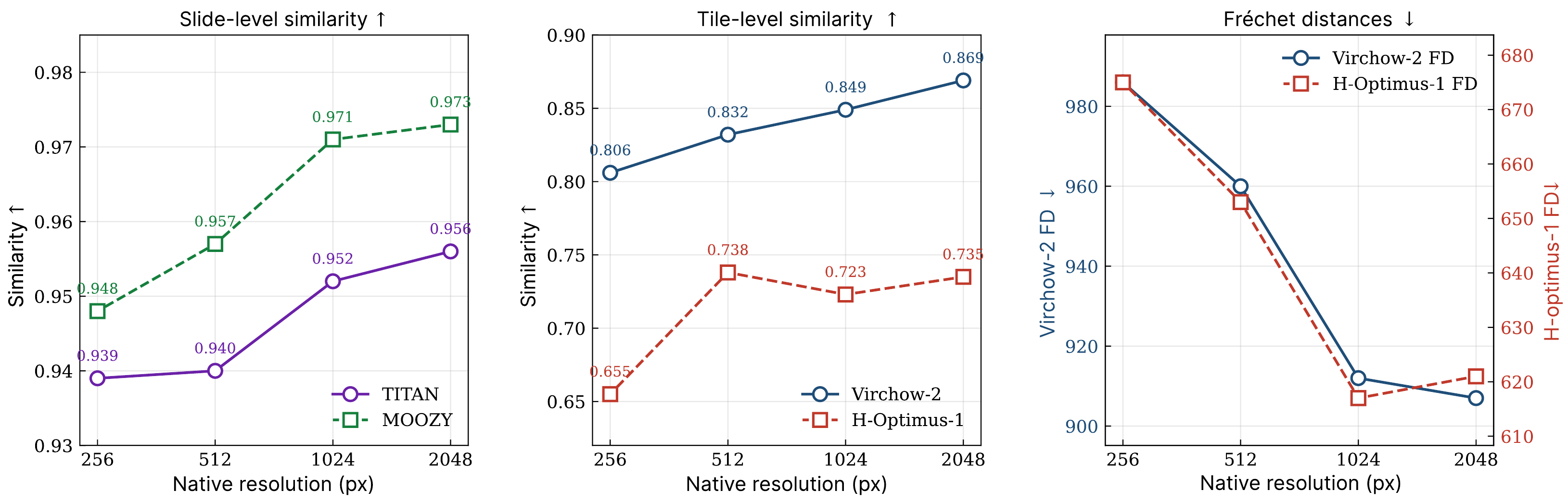}
    \caption{Effect of tile generator's native resolution on 8192\ttimes8192 generation quality. Both slide-level and tile-level metrics improve with resolution, with diminishing returns beyond 1024.}
    \label{fig:resolutions_quality}
\end{figure}

We train \methodtile at four native resolutions as a part of our progressive training pipeline (256, 512, 1024, 2048). Here we evaluate each model checkpoint at 8192\ttimes8192 generation using the sliding-window pipeline. As shown in Figure \ref{fig:resolutions_quality}, both slide-level and tile-level metrics improve monotonically with native resolution. Most of the improvement occurs below 1024 px, with the 1024 $\rightarrow$ 2048 step yielding smaller gains.

Per-sample training cost scales by ~8\ttimes from 1024 to 2048 (Table \ref{tab:resolutions_time}), reflecting the quadratic attention cost in latent sequence length. A single-iteration benchmark at 4096 pixels measures a further 12\ttimes increase, putting full 4096 training at $\sim$ 96 times   our 1024\ttimes1024 budget, which is impractical at the dataset scale used in this paper. Combined with the diminishing fidelity gains beyond 1024, this leads us to adopt 2048 as the final operating point.

\subsection{Resolution-matched comparisons}
\label{ablation:matched}

Table~\ref{tab:8k_baselines} compares tile generators trained at different native resolutions. Since fidelity increases with native resolution (Appendix~\ref{ablation:tile_resolution}), that comparison conflates model design with training resolution. We disentangle the two with two additional comparisons at 8192\ttimes8192 on the same 100 held-out ROIs (Table~\ref{tab:resolution_matched}).

\noindent\textbf{Matched native resolution.} Our progressive training strategy produces a 1024-native \method checkpoint (Appendix~\ref{ablation:tile_resolution}), which we compare against PixCell-1024 \cite{pixcell}. At identical native resolution and identical inference, \method leads on every slide- and patch-level metric.

\noindent\textbf{Matched native resolution, data, and compute.} We train a 2048-native model that keeps our backbone (SiT-XL, Flux-2 VAE latent space), training data, and sampler fixed, and replaces our dense embedding conditioning with the summarizer-based conditioning of ZoomLDM \cite{zoomldm}, under the same training budget as \method. Following ZoomLDM, we train on fixed-size tiles at five magnifications (20\ttimes{} to 1.25\ttimes{}) and use its joint sampling algorithm to synthesize 8192\ttimes8192 images at 20\ttimes{}, guided by a 2048\ttimes2048 image at 5\ttimes{}. The summarizer-conditioned model (\method-summarizer) underperforms not only \method but also PixCell-1024 on every metric. We attribute this to ZoomLDM's pooling of the conditioning grid to a fixed $64 \times 1536$ budget: at 2048 native resolution, a 20\ttimes{} tile (a 2048\ttimes2048 region), a 5\ttimes{} tile (an 8192\ttimes8192 region), and a 1.25\ttimes{} tile (a 32,768\ttimes32,768 region) are all conditioned with the same $64 \times 1536$ grid. We believe this is the bottleneck when scaling ZoomLDM in resolution, and it warrants further dedicated investigation.

\begin{table}[h]
    \centering
    \caption{Resolution-matched comparisons at 8192\ttimes8192 resolution ($N=100$, same ROIs as Table~\ref{tab:8k_baselines}). Top: \method and PixCell at the same 1024 native resolution. Bottom: \method against a model with identical backbone, data, sampler, and training budget that uses ZoomLDM's summarizer-based conditioning.}
    \vspace{1em}
    \begin{tabular}{lcccccc}
    \toprule
    & \multicolumn{2}{c}{Slide-level Similarity} & \multicolumn{2}{c}{Patch-level Similarity} & \multicolumn{2}{c}{Fréchet Distance} \\
    \cmidrule(lr){2-3} \cmidrule(lr){4-5} \cmidrule(lr){6-7}
    Generator & TITAN\textuparrow & MOOZY\textuparrow & V2\textsubscript{mean}\textuparrow & H1\textsubscript{mean}\textuparrow & V2\textsubscript{FD}\textdownarrow & H1\textsubscript{FD}\textdownarrow \\
    \midrule
    \multicolumn{7}{l}{\textit{Matched native resolution}} \\
    PixCell-1024               & .941          & .900          & .778          & .667          & 1023          & 696          \\
    \method-1024               & \textbf{.952} & \textbf{.971} & \textbf{.849} & \textbf{.723} & \textbf{912}  & \textbf{617} \\
    \midrule
    \multicolumn{7}{l}{\textit{Matched native resolution, data, and compute}} \\
    \method-summarizer-2048    & .926          & .868          & .627          & .505          & 1229          & 813          \\
    \method-2048               & \textbf{.956} & \textbf{.973} & \textbf{.869} & \textbf{.735} & \textbf{907}  & \textbf{621} \\
    \bottomrule
\end{tabular}

    \label{tab:resolution_matched}
\end{table}

\begin{figure}
    \centering
    \includegraphics[width=\linewidth]{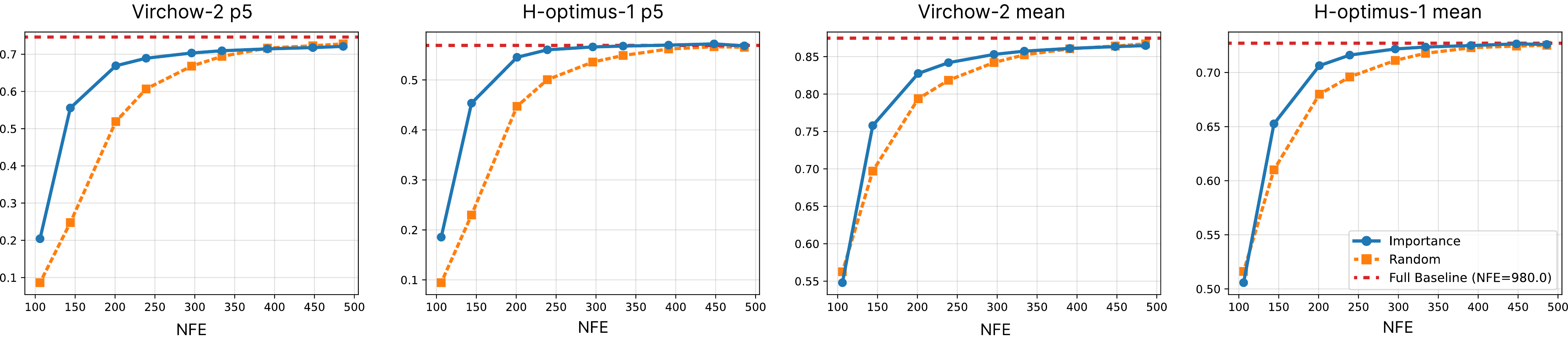}
    \caption{Effect of window selection strategy on 8192\ttimes8192 generation quality. Our CFG importance-based strategy outperforms random selection across all metrics.}
    \label{fig:8k_plots_ablation}
\end{figure}

\subsection{Importance-based window selection}

We isolate the contribution of the importance signal by comparing our top-$r$ CFG-based selection against a random-$r$ baseline that selects the same fraction of windows uniformly at random. Both schedules use the same number of denoising steps and velocity caching; the only difference is which windows get fresh vs. cached velocities. We sweep $r$ to vary NFE and evaluate on 8192\ttimes8192 generation.

As shown in Figure \ref{fig:8k_plots_ablation}, importance-based selection outperforms random selection across the full NFE range, with larger gaps on tail metrics at low compute. The gap closes as NFE grows and both schedules approach the full-compute ceiling. This confirms that the CFG magnitude signal identifies windows that genuinely benefit from fresh velocity computation, while random selection is more likely to waste compute on unimportant windows.

\section{Report-level agreement with MedGemma}
\label{appendix:medgemma}

To assess generated images without relying on pathology foundation model embeddings, we process each generated image and its reference with MedGemma-4B \cite{medgemma}, a general-purpose medical vision-language model, and compare the decoded text. Patching follows the high-dimensional pathology notebook released with the model. Each image is presented in a single turn with a fixed template that the model must fill:

\begin{verbatim}
    TISSUE OF ORIGIN: <one organ>
    MALIGNANT: <yes or no>
    DIAGNOSIS: <one most likely specific diagnosis>
\end{verbatim}

We report the fraction of generated/reference pairs whose reports agree on the tissue of origin, on the diagnosis, and on all three fields. The 8192\ttimes8192 comparison across methods is shown in Table~\ref{tab:medgemma_8k}.

\noindent\textbf{Gigapixel generation.} We repeat this evaluation on the 256 matched real/synthetic 32,768\ttimes32,768 pairs used in Section~\ref{sec:result_efficient_generation}, generated at full compute ($r=1$). As a chance baseline, we compute the rate at which a real report matches the synthetic report of a different slide, averaged over all such pairings. Table~\ref{tab:medgemma_32k} shows that agreement between \method generations and their reference slides far exceeds chance on all three criteria, indicating that our 32k generations carry the diagnostic content of the tissue they are conditioned on.

\begin{table}[h]
    \centering
    \caption{Report-level agreement at 32,768\ttimes32,768 resolution ($N=256$, full compute). Chance is the rate at which a real report matches the synthetic report of a different slide, averaged over all such pairings. 95\% Wilson intervals in brackets.}
    \vspace{1em}
    \begin{tabular}{lcc}
    \toprule
    Field agreement & \method vs.\ reference & Chance \\
    \midrule
    Same tissue of origin      & 85.9\% [81.1, 89.6] & 9.1\% [7.7, 10.7] \\
    Same diagnosis             & 72.2\% [66.4, 77.3] & 9.2\% [7.4, 11.1] \\
    All three fields identical & 71.0\% [65.1, 76.2] & 6.0\% [4.6, 7.7]  \\
    \bottomrule
\end{tabular}

    \label{tab:medgemma_32k}
\end{table}

\section{LLM usage}
\label{llm_usage}

\subsection{Report summarization for CGM training}

Pathology reports in TCGA and HistAI are unstructured free-form text. To produce concise summaries suitable for conditioning the CGM, we prompt the Qwen-3-4B-Instruct \cite{qwen3} model to rewrite each report into a concise summary, capturing key diagnostic information. We generate five paraphrases per report to provide diversity during CGM training. The HistAI dataset \cite{histai} includes reports along with the images. For TCGA, we use cleaned reports from  \citet{kefeli2024tcga}. We use the following prompt template:

\begin{verbatim}
    You are a pathology lab assistant. You are given an unstructured pathology
report describing a tissue sample. Follow these instructions carefully and
to the letter:
- Extract a detailed summary of the diagnosis and the examined tissue from
  the report in a sentence under 50 words.
- Mention tissue type if it is mentioned in the report.
- If the report includes results of immunohistochemical studies or molecular
  tests, include a short summary of the most important positive results.
  Keep the overall sentence length below 50 words.
- Do not mention any negative test results or absence of something.
- Minimize prose, be succinct, use as few words as possible.
- Following the instructions above, rewrite the report 5 times, for a total
  of 5 sentences, each with a summary written in a slightly different way.
  Make sure all summaries are consistent with each other.
- Output the resulting 5 sentences as a list in JSON format. Don't output
  anything else.
\end{verbatim}

\subsection{Synthetic report generation for data augmentation}

The CGM-based augmentation procedure of Section \ref{sec:cdm_augmentation} requires a class-conditional text prompt for each label. For both TCGA-UT-8K (32 cancer types) and PANDA (6 ISUP grades), we use Claude Opus 4.7 to generate four short plausible pathology reports per class, following this prompt template:

\begin{verbatim}
    I want to perform a class-conditional augmentation task.
The task is to predict <label name> (<label range or class list>). For each
label, give me four plausible pathology reports. The report should be no
more than 25 words. Return the output in json.
\end{verbatim}

\begin{figure}
    \centering
    \includegraphics[width=\linewidth]{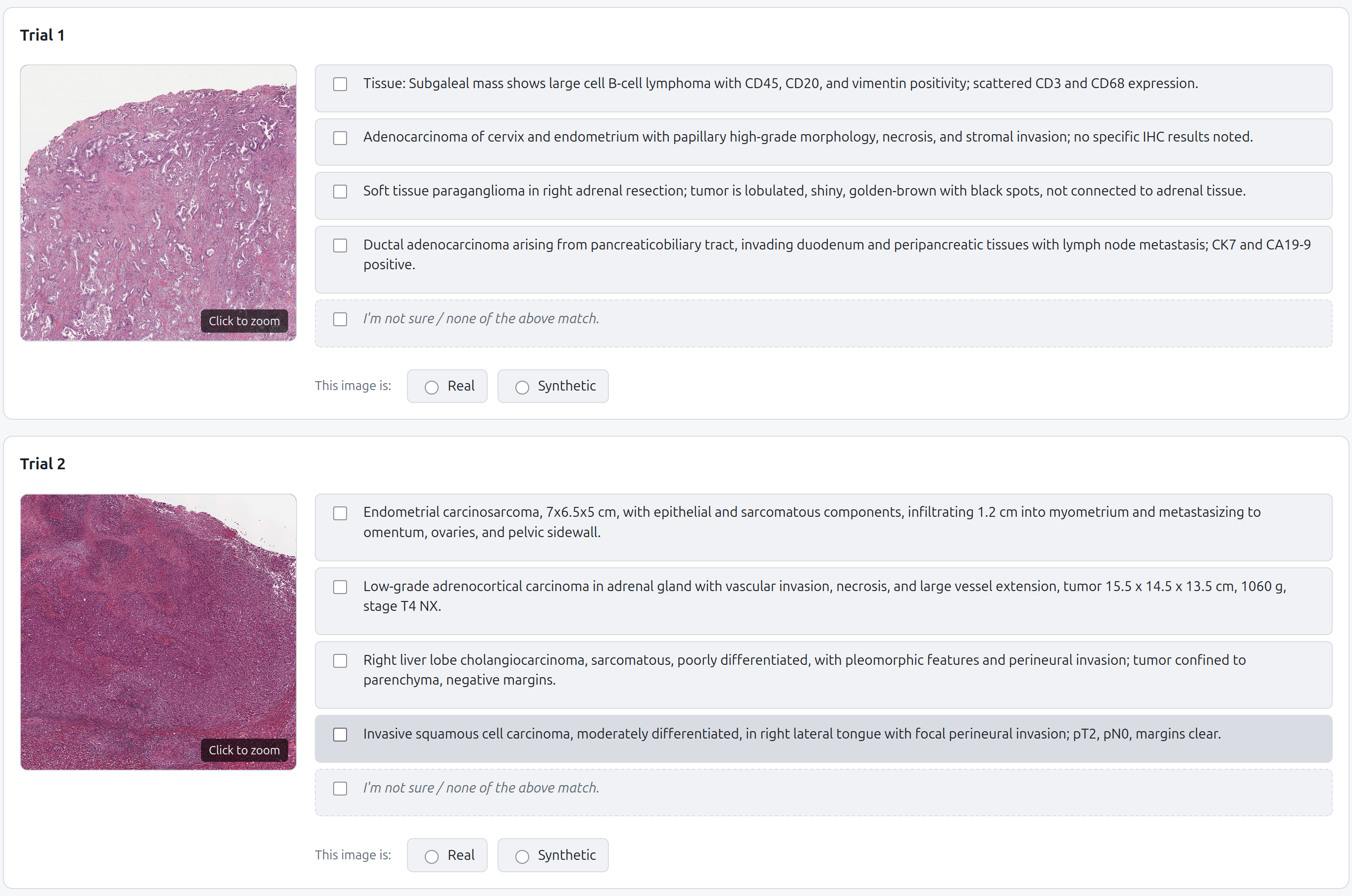}
    \caption{Pathologist evaluation interface. Each trial shows an 8192\ttimes8192 image (real or synthetic) and four candidate reports (one of which is true). Pathologists indicate matching report(s) and classify the image as real or synthetic.}
    \label{fig:eval_study}
\end{figure}

\section{Broader impacts}
\label{broader_impacts}

\method enables high-quality generation of gigapixel histopathology images, which can be a powerful tool for research and education. As it is trained on publicly available data, the model reflects the biases present in those datasets. Generated images are intended for research use only and should not be used as a substitute for real tissue in clinical or diagnostic settings.

\begin{figure}
    \centering
    \includegraphics[width=\linewidth]{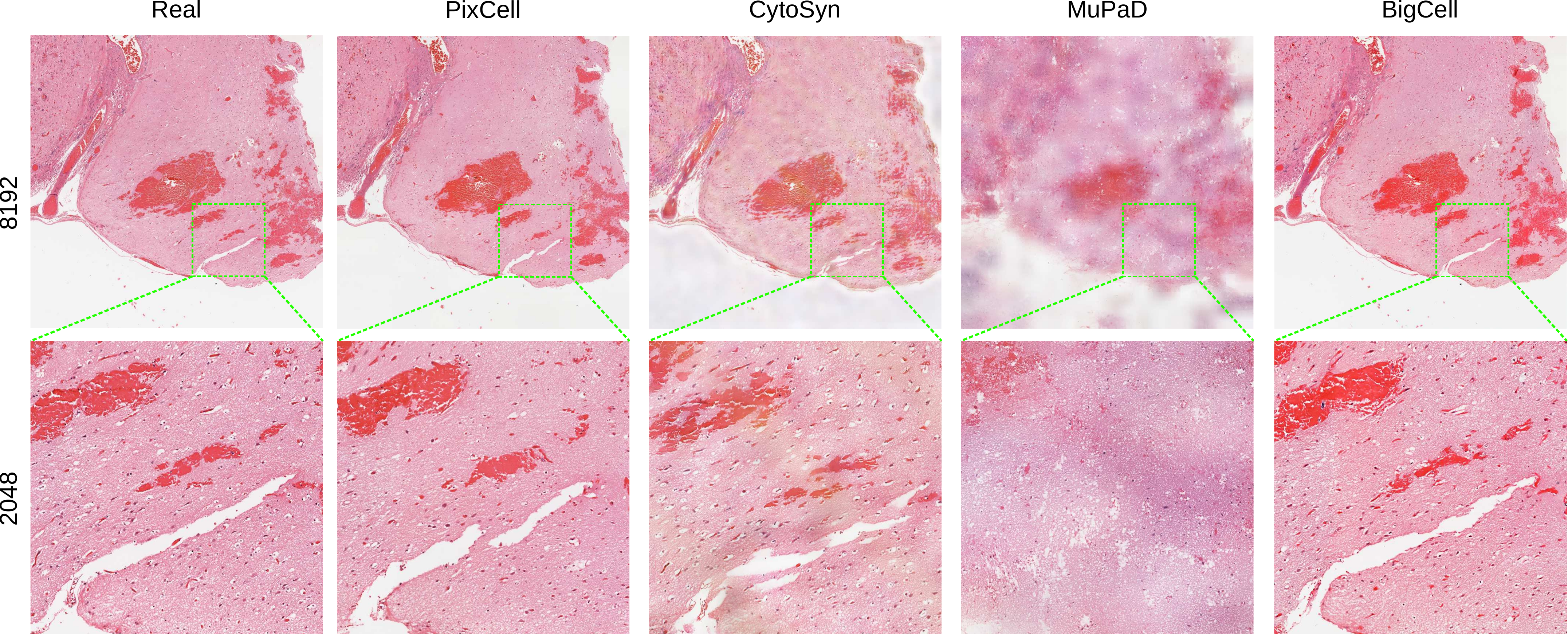}
    \caption{Qualitative comparison of 8192\ttimes8192 generation between \methodtile  and baselines. Top row: full 8k image. Bottom row: 2k crop from the boxed region.}
    \label{fig:baselines_8k}
\end{figure}

\section{Additional qualitative results}
\label{appendix_figures}

Figure \ref{fig:baselines_8k} shows 8192\ttimes8192 samples from \methodtile and baseline tile generators, along with the reference real image. \methodtile preserves both the global tissue structure and fine cellular details faithfully. PixCell preserves local details but generates structures that are inconsistent with the reference. CytoSyn captures the overall structure but exhibits blurriness, consistent with its lower similarity metrics. MuPaD struggles at this resolution -- the global layout collapses and tissue morphology is poorly preserved, in line with its substantially worse metrics. 

\begin{figure}
    \centering
    \includegraphics[width=\linewidth]{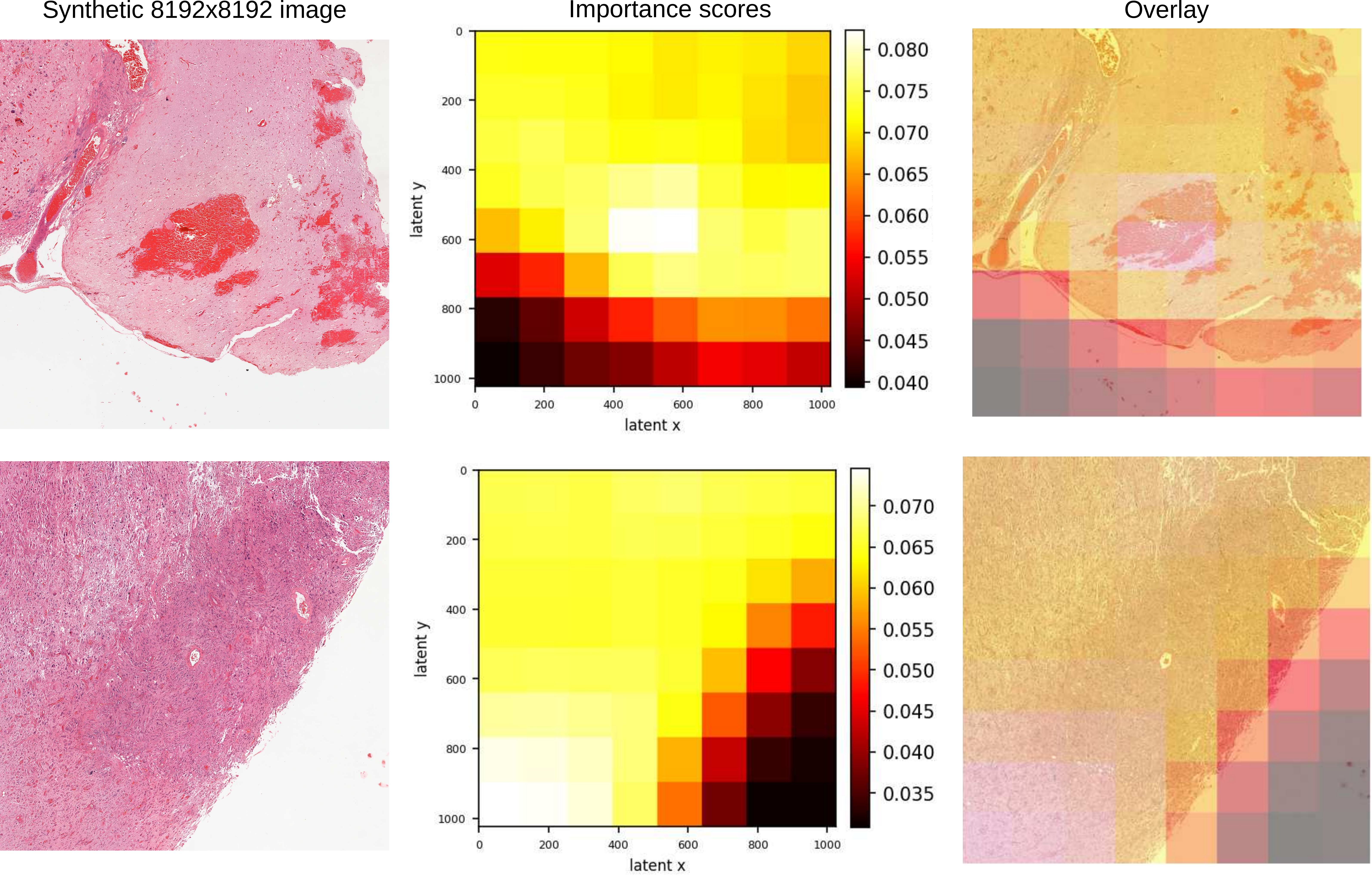}
    \caption{Visualization of CFG-magnitude importance scores on two synthetic 8192\ttimes8192 images. Tissue regions receive high importance, while background regions receive low importance.}
    \label{fig:cfg_magnitude}
\end{figure}

In Figure~\ref{fig:cfg_magnitude}, we visualize the importance assigned to different windows in an image during the efficient sliding-window generation. As expected, the model adaptively selects the regions with more details as the ones to perform more denoising steps for, while leaving background areas to be generated with fewer model forward passes. This speeds up generation significantly without impacting the overall quality of the gigapixel image produced.

\begin{figure}
    \centering
    \includegraphics[width=\linewidth]{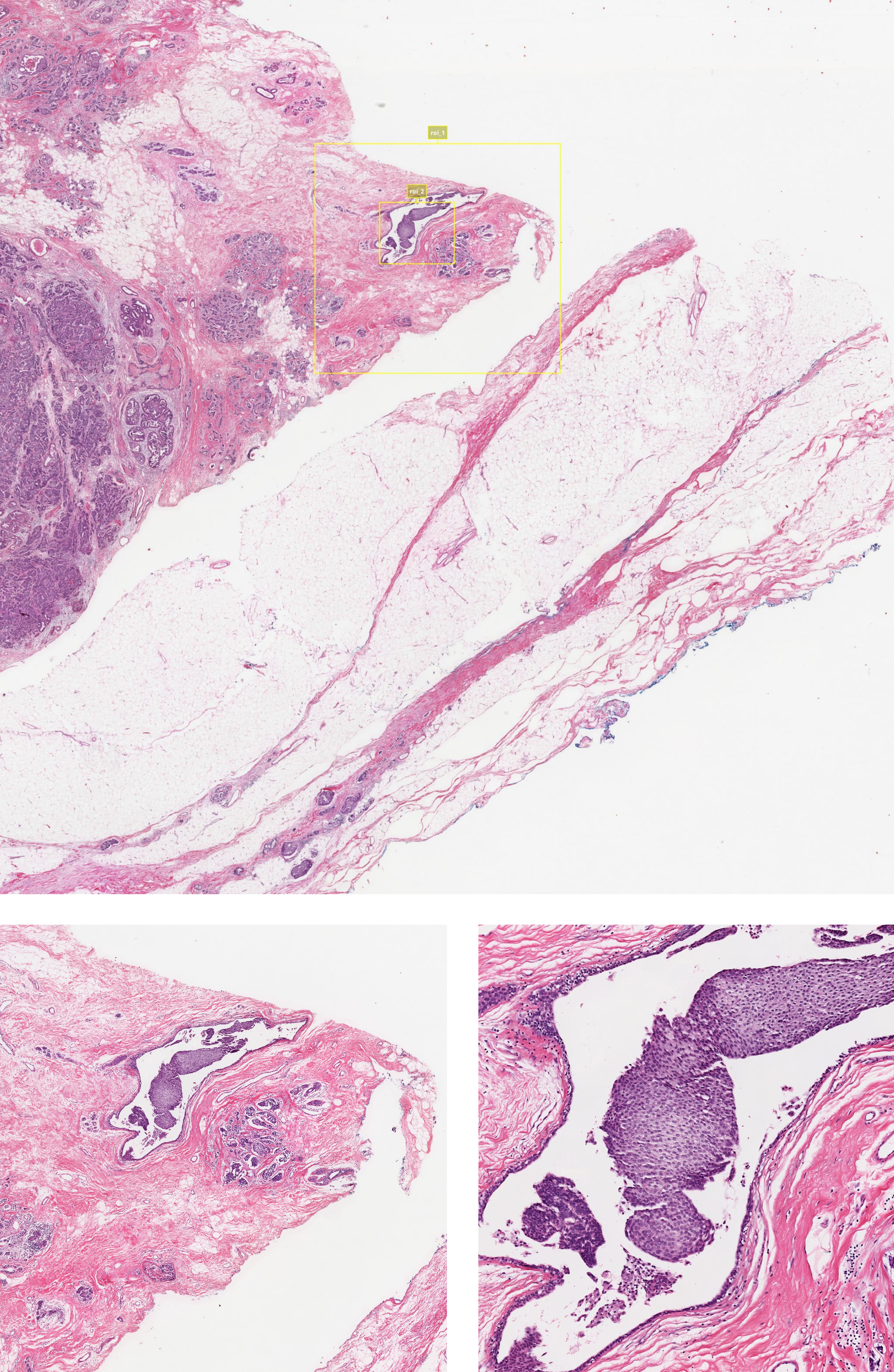}
    \caption{32,768\ttimes32,768 pixel breast tissue image generated by \method (top), with progressive zoom-in crops from the marked regions (bottom).}
    \label{fig:32k_brca}
\end{figure}

\begin{figure}
    \centering
    \includegraphics[width=\linewidth]{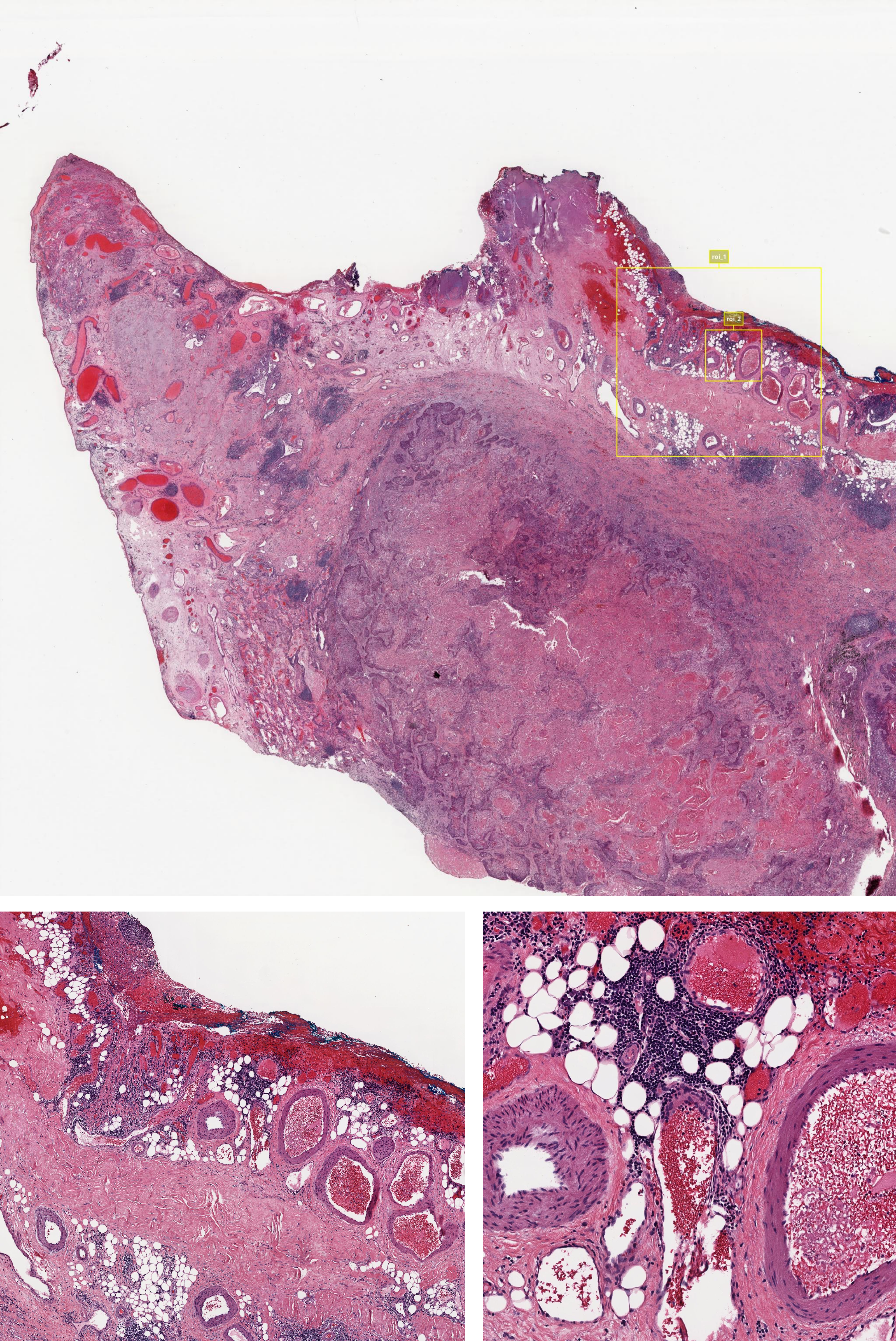}
    \caption{32,768\ttimes32,768 pixel lung tissue image generated by \method (top), with progressive zoom-in crops from the marked regions (bottom).}
    \label{fig:32k_lusc}
\end{figure}

\begin{figure}
    \centering
    \includegraphics[width=\linewidth]{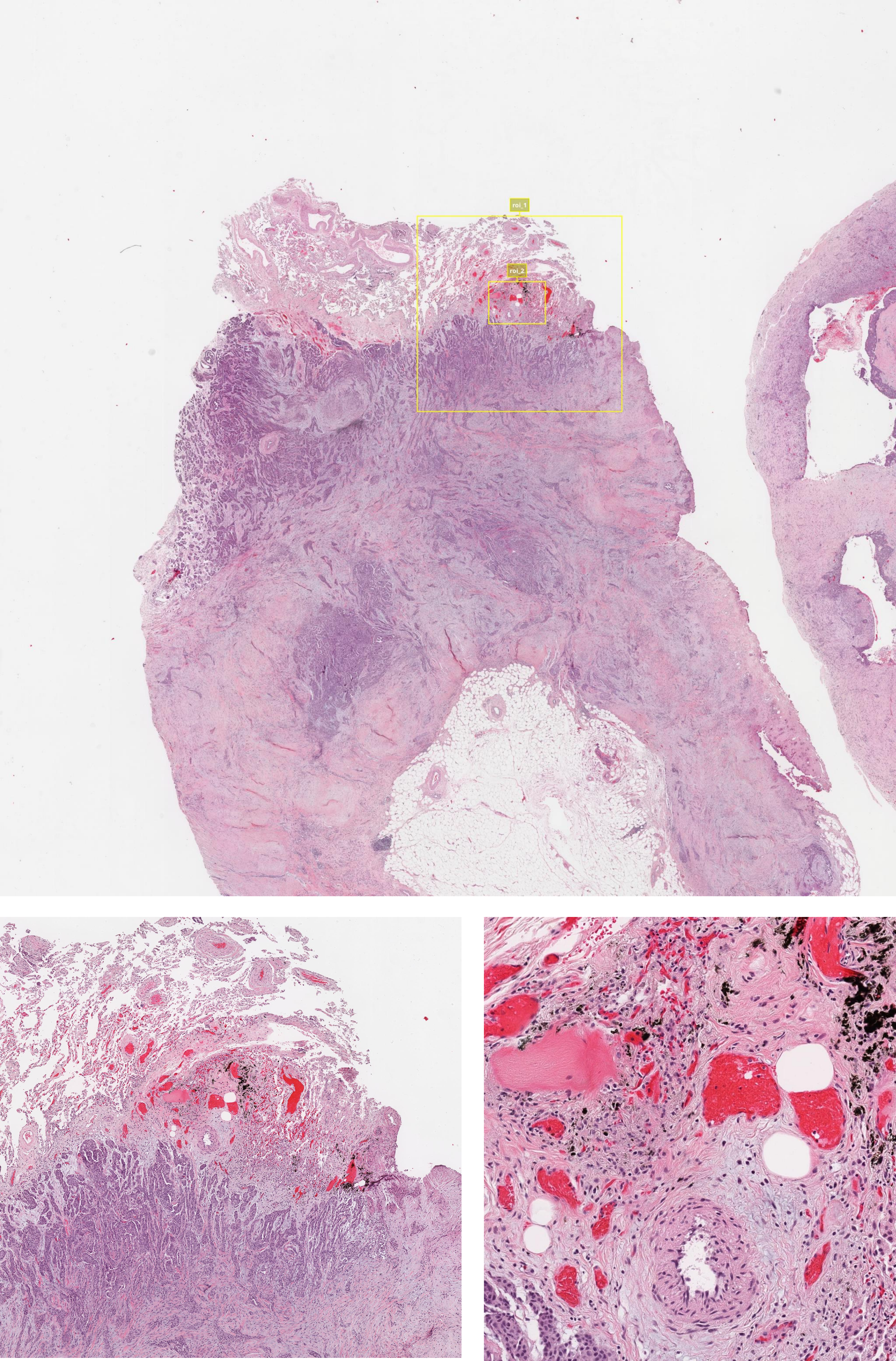}
    \caption{32,768\ttimes32,768 pixel mesothelial tissue image generated by \method (top), with progressive zoom-in crops from the marked regions (bottom).}
    \label{fig:32k_meso}
\end{figure}

In Figures \ref{fig:32k_brca}, \ref{fig:32k_lusc}, \ref{fig:32k_meso}, we show synthetic 32,768\ttimes32,768 pixel images generated by \method. Full-resolution pyramidal TIFFs of additional synthetic 32,768\ttimes32,768 samples are available at \url{https://figshare.com/s/f52821e239de442b7967} and can be viewed interactively using the QuPath\footnote{\url{https://qupath.github.io}} software.

\newpage
\section*{NeurIPS Paper Checklist}

\begin{enumerate}

\item {\bf Claims}
    \item[] Question: Do the main claims made in the abstract and introduction accurately reflect the paper's contributions and scope?
    \item[] Answer: \answerYes{} 
    \item[] Justification: The results shown in the paper reflect the contributions described in the abstract and introduction. 
    \item[] Guidelines:
    \begin{itemize}
        \item The answer \answerNA{} means that the abstract and introduction do not include the claims made in the paper.
        \item The abstract and/or introduction should clearly state the claims made, including the contributions made in the paper and important assumptions and limitations. A \answerNo{} or \answerNA{} answer to this question will not be perceived well by the reviewers. 
        \item The claims made should match theoretical and experimental results, and reflect how much the results can be expected to generalize to other settings. 
        \item It is fine to include aspirational goals as motivation as long as it is clear that these goals are not attained by the paper. 
    \end{itemize}

\item {\bf Limitations}
    \item[] Question: Does the paper discuss the limitations of the work performed by the authors?
    \item[] Answer: \answerYes{} 
    \item[] Justification: We discuss the limitations of our work in Section~\ref{sec:limitations}.
    \item[] Guidelines:
    \begin{itemize}
        \item The answer \answerNA{} means that the paper has no limitation while the answer \answerNo{} means that the paper has limitations, but those are not discussed in the paper. 
        \item The authors are encouraged to create a separate ``Limitations'' section in their paper.
        \item The paper should point out any strong assumptions and how robust the results are to violations of these assumptions (e.g., independence assumptions, noiseless settings, model well-specification, asymptotic approximations only holding locally). The authors should reflect on how these assumptions might be violated in practice and what the implications would be.
        \item The authors should reflect on the scope of the claims made, e.g., if the approach was only tested on a few datasets or with a few runs. In general, empirical results often depend on implicit assumptions, which should be articulated.
        \item The authors should reflect on the factors that influence the performance of the approach. For example, a facial recognition algorithm may perform poorly when image resolution is low or images are taken in low lighting. Or a speech-to-text system might not be used reliably to provide closed captions for online lectures because it fails to handle technical jargon.
        \item The authors should discuss the computational efficiency of the proposed algorithms and how they scale with dataset size.
        \item If applicable, the authors should discuss possible limitations of their approach to address problems of privacy and fairness.
        \item While the authors might fear that complete honesty about limitations might be used by reviewers as grounds for rejection, a worse outcome might be that reviewers discover limitations that aren't acknowledged in the paper. The authors should use their best judgment and recognize that individual actions in favor of transparency play an important role in developing norms that preserve the integrity of the community. Reviewers will be specifically instructed to not penalize honesty concerning limitations.
    \end{itemize}

\item {\bf Theory assumptions and proofs}
    \item[] Question: For each theoretical result, does the paper provide the full set of assumptions and a complete (and correct) proof?
    \item[] Answer: \answerNA{} 
    \item[] Justification: We do not include theoretical results in this paper.
    \item[] Guidelines:
    \begin{itemize}
        \item The answer \answerNA{} means that the paper does not include theoretical results. 
        \item All the theorems, formulas, and proofs in the paper should be numbered and cross-referenced.
        \item All assumptions should be clearly stated or referenced in the statement of any theorems.
        \item The proofs can either appear in the main paper or the supplemental material, but if they appear in the supplemental material, the authors are encouraged to provide a short proof sketch to provide intuition. 
        \item Inversely, any informal proof provided in the core of the paper should be complemented by formal proofs provided in appendix or supplemental material.
        \item Theorems and Lemmas that the proof relies upon should be properly referenced. 
    \end{itemize}

    \item {\bf Experimental result reproducibility}
    \item[] Question: Does the paper fully disclose all the information needed to reproduce the main experimental results of the paper to the extent that it affects the main claims and/or conclusions of the paper (regardless of whether the code and data are provided or not)?
    \item[] Answer: \answerYes{} 
    \item[] Justification: We provide all the information needed to reproduce the main experimental results in Appendix Section~\ref{appendix:implementation_details}.
    \item[] Guidelines:
    \begin{itemize}
        \item The answer \answerNA{} means that the paper does not include experiments.
        \item If the paper includes experiments, a \answerNo{} answer to this question will not be perceived well by the reviewers: Making the paper reproducible is important, regardless of whether the code and data are provided or not.
        \item If the contribution is a dataset and\slash or model, the authors should describe the steps taken to make their results reproducible or verifiable. 
        \item Depending on the contribution, reproducibility can be accomplished in various ways. For example, if the contribution is a novel architecture, describing the architecture fully might suffice, or if the contribution is a specific model and empirical evaluation, it may be necessary to either make it possible for others to replicate the model with the same dataset, or provide access to the model. In general. releasing code and data is often one good way to accomplish this, but reproducibility can also be provided via detailed instructions for how to replicate the results, access to a hosted model (e.g., in the case of a large language model), releasing of a model checkpoint, or other means that are appropriate to the research performed.
        \item While NeurIPS does not require releasing code, the conference does require all submissions to provide some reasonable avenue for reproducibility, which may depend on the nature of the contribution. For example
        \begin{enumerate}
            \item If the contribution is primarily a new algorithm, the paper should make it clear how to reproduce that algorithm.
            \item If the contribution is primarily a new model architecture, the paper should describe the architecture clearly and fully.
            \item If the contribution is a new model (e.g., a large language model), then there should either be a way to access this model for reproducing the results or a way to reproduce the model (e.g., with an open-source dataset or instructions for how to construct the dataset).
            \item We recognize that reproducibility may be tricky in some cases, in which case authors are welcome to describe the particular way they provide for reproducibility. In the case of closed-source models, it may be that access to the model is limited in some way (e.g., to registered users), but it should be possible for other researchers to have some path to reproducing or verifying the results.
        \end{enumerate}
    \end{itemize}

\item {\bf Open access to data and code}
    \item[] Question: Does the paper provide open access to the data and code, with sufficient instructions to faithfully reproduce the main experimental results, as described in supplemental material?
    \item[] Answer: \answerYes{} 
    \item[] Justification: The entire \method pipeline is trained exclusively on publicly available datasets with no proprietary data. We open source our code and pretrained models at this \href{https://histodiffusion.github.io/docs/projects/bigcell/}{link}.
    \item[] Guidelines:
    \begin{itemize}
        \item The answer \answerNA{} means that paper does not include experiments requiring code.
        \item Please see the NeurIPS code and data submission guidelines (\url{https://neurips.cc/public/guides/CodeSubmissionPolicy}) for more details.
        \item While we encourage the release of code and data, we understand that this might not be possible, so \answerNo{} is an acceptable answer. Papers cannot be rejected simply for not including code, unless this is central to the contribution (e.g., for a new open-source benchmark).
        \item The instructions should contain the exact command and environment needed to run to reproduce the results. See the NeurIPS code and data submission guidelines (\url{https://neurips.cc/public/guides/CodeSubmissionPolicy}) for more details.
        \item The authors should provide instructions on data access and preparation, including how to access the raw data, preprocessed data, intermediate data, and generated data, etc.
        \item The authors should provide scripts to reproduce all experimental results for the new proposed method and baselines. If only a subset of experiments are reproducible, they should state which ones are omitted from the script and why.
        \item At submission time, to preserve anonymity, the authors should release anonymized versions (if applicable).
        \item Providing as much information as possible in supplemental material (appended to the paper) is recommended, but including URLs to data and code is permitted.
    \end{itemize}

\item {\bf Experimental setting/details}
    \item[] Question: Does the paper specify all the training and test details (e.g., data splits, hyperparameters, how they were chosen, type of optimizer) necessary to understand the results?
    \item[] Answer: \answerYes{} 
    \item[] Justification: Please refer to Section~\ref{sec:experimental_setup} and Appendix~\ref{appendix:implementation_details} for all the details of our experimental setting.
    \item[] Guidelines:
    \begin{itemize}
        \item The answer \answerNA{} means that the paper does not include experiments.
        \item The experimental setting should be presented in the core of the paper to a level of detail that is necessary to appreciate the results and make sense of them.
        \item The full details can be provided either with the code, in appendix, or as supplemental material.
    \end{itemize}

\item {\bf Experiment statistical significance}
    \item[] Question: Does the paper report error bars suitably and correctly defined or other appropriate information about the statistical significance of the experiments?
    \item[] Answer: \answerYes{} 
    \item[] Justification: We perform statistical significance tests for the data augmentation experiments and comparison against prior tile generators. 
    \item[] Guidelines:
    \begin{itemize}
        \item The answer \answerNA{} means that the paper does not include experiments.
        \item The authors should answer \answerYes{} if the results are accompanied by error bars, confidence intervals, or statistical significance tests, at least for the experiments that support the main claims of the paper.
        \item The factors of variability that the error bars are capturing should be clearly stated (for example, train/test split, initialization, random drawing of some parameter, or overall run with given experimental conditions).
        \item The method for calculating the error bars should be explained (closed form formula, call to a library function, bootstrap, etc.)
        \item The assumptions made should be given (e.g., Normally distributed errors).
        \item It should be clear whether the error bar is the standard deviation or the standard error of the mean.
        \item It is OK to report 1-sigma error bars, but one should state it. The authors should preferably report a 2-sigma error bar than state that they have a 96\% CI, if the hypothesis of Normality of errors is not verified.
        \item For asymmetric distributions, the authors should be careful not to show in tables or figures symmetric error bars that would yield results that are out of range (e.g., negative error rates).
        \item If error bars are reported in tables or plots, the authors should explain in the text how they were calculated and reference the corresponding figures or tables in the text.
    \end{itemize}

\item {\bf Experiments compute resources}
    \item[] Question: For each experiment, does the paper provide sufficient information on the computer resources (type of compute workers, memory, time of execution) needed to reproduce the experiments?
    \item[] Answer: \answerYes{} 
    \item[] Justification: Please refer to Appendix Section~\ref{appendix:implementation_details}.
    \item[] Guidelines:
    \begin{itemize}
        \item The answer \answerNA{} means that the paper does not include experiments.
        \item The paper should indicate the type of compute workers CPU or GPU, internal cluster, or cloud provider, including relevant memory and storage.
        \item The paper should provide the amount of compute required for each of the individual experimental runs as well as estimate the total compute. 
        \item The paper should disclose whether the full research project required more compute than the experiments reported in the paper (e.g., preliminary or failed experiments that didn't make it into the paper). 
    \end{itemize}
    
\item {\bf Code of ethics}
    \item[] Question: Does the research conducted in the paper conform, in every respect, with the NeurIPS Code of Ethics \url{https://neurips.cc/public/EthicsGuidelines}?
    \item[] Answer: \answerYes{} 
    \item[] Justification: We follow the NeurIPS Code of Ethics.
    \item[] Guidelines:
    \begin{itemize}
        \item The answer \answerNA{} means that the authors have not reviewed the NeurIPS Code of Ethics.
        \item If the authors answer \answerNo, they should explain the special circumstances that require a deviation from the Code of Ethics.
        \item The authors should make sure to preserve anonymity (e.g., if there is a special consideration due to laws or regulations in their jurisdiction).
    \end{itemize}

\item {\bf Broader impacts}
    \item[] Question: Does the paper discuss both potential positive societal impacts and negative societal impacts of the work performed?
    \item[] Answer: \answerYes{} 
    \item[] Justification: We discuss this in Appendix Section~\ref{broader_impacts}.
    \item[] Guidelines:
    \begin{itemize}
        \item The answer \answerNA{} means that there is no societal impact of the work performed.
        \item If the authors answer \answerNA{} or \answerNo, they should explain why their work has no societal impact or why the paper does not address societal impact.
        \item Examples of negative societal impacts include potential malicious or unintended uses (e.g., disinformation, generating fake profiles, surveillance), fairness considerations (e.g., deployment of technologies that could make decisions that unfairly impact specific groups), privacy considerations, and security considerations.
        \item The conference expects that many papers will be foundational research and not tied to particular applications, let alone deployments. However, if there is a direct path to any negative applications, the authors should point it out. For example, it is legitimate to point out that an improvement in the quality of generative models could be used to generate Deepfakes for disinformation. On the other hand, it is not needed to point out that a generic algorithm for optimizing neural networks could enable people to train models that generate Deepfakes faster.
        \item The authors should consider possible harms that could arise when the technology is being used as intended and functioning correctly, harms that could arise when the technology is being used as intended but gives incorrect results, and harms following from (intentional or unintentional) misuse of the technology.
        \item If there are negative societal impacts, the authors could also discuss possible mitigation strategies (e.g., gated release of models, providing defenses in addition to attacks, mechanisms for monitoring misuse, mechanisms to monitor how a system learns from feedback over time, improving the efficiency and accessibility of ML).
    \end{itemize}
    
\item {\bf Safeguards}
    \item[] Question: Does the paper describe safeguards that have been put in place for responsible release of data or models that have a high risk for misuse (e.g., pre-trained language models, image generators, or scraped datasets)?
    \item[] Answer: \answerYes{} 
    \item[] Justification: Users of our released models and datasets will be required to adhere to usage guidelines and restrictions to access the model.
    \item[] Guidelines:
    \begin{itemize}
        \item The answer \answerNA{} means that the paper poses no such risks.
        \item Released models that have a high risk for misuse or dual-use should be released with necessary safeguards to allow for controlled use of the model, for example by requiring that users adhere to usage guidelines or restrictions to access the model or implementing safety filters. 
        \item Datasets that have been scraped from the Internet could pose safety risks. The authors should describe how they avoided releasing unsafe images.
        \item We recognize that providing effective safeguards is challenging, and many papers do not require this, but we encourage authors to take this into account and make a best faith effort.
    \end{itemize}

\item {\bf Licenses for existing assets}
    \item[] Question: Are the creators or original owners of assets (e.g., code, data, models), used in the paper, properly credited and are the license and terms of use explicitly mentioned and properly respected?
    \item[] Answer: \answerYes{} 
    \item[] Justification: We cite all the original assets used in our paper.
    \item[] Guidelines:
    \begin{itemize}
        \item The answer \answerNA{} means that the paper does not use existing assets.
        \item The authors should cite the original paper that produced the code package or dataset.
        \item The authors should state which version of the asset is used and, if possible, include a URL.
        \item The name of the license (e.g., CC-BY 4.0) should be included for each asset.
        \item For scraped data from a particular source (e.g., website), the copyright and terms of service of that source should be provided.
        \item If assets are released, the license, copyright information, and terms of use in the package should be provided. For popular datasets, \url{paperswithcode.com/datasets} has curated licenses for some datasets. Their licensing guide can help determine the license of a dataset.
        \item For existing datasets that are re-packaged, both the original license and the license of the derived asset (if it has changed) should be provided.
        \item If this information is not available online, the authors are encouraged to reach out to the asset's creators.
    \end{itemize}

\item {\bf New assets}
    \item[] Question: Are new assets introduced in the paper well documented and is the documentation provided alongside the assets?
    \item[] Answer: \answerYes{} 
    \item[] Justification:  We open-source our code and pre-trained weights at this \href{https://histodiffusion.github.io/docs/projects/bigcell/}{link}. 
    \item[] Guidelines:
    \begin{itemize}
        \item The answer \answerNA{} means that the paper does not release new assets.
        \item Researchers should communicate the details of the dataset\slash code\slash model as part of their submissions via structured templates. This includes details about training, license, limitations, etc. 
        \item The paper should discuss whether and how consent was obtained from people whose asset is used.
        \item At submission time, remember to anonymize your assets (if applicable). You can either create an anonymized URL or include an anonymized zip file.
    \end{itemize}

\item {\bf Crowdsourcing and research with human subjects}
    \item[] Question: For crowdsourcing experiments and research with human subjects, does the paper include the full text of instructions given to participants and screenshots, if applicable, as well as details about compensation (if any)? 
    \item[] Answer: \answerYes{} 
    \item[] Justification: We ask expert pathologists to evaluate the synthetic images. We provide a screenshot of the evaluation platform in Figure \ref{fig:eval_study}.
    \item[] Guidelines:
    \begin{itemize}
        \item The answer \answerNA{} means that the paper does not involve crowdsourcing nor research with human subjects.
        \item Including this information in the supplemental material is fine, but if the main contribution of the paper involves human subjects, then as much detail as possible should be included in the main paper. 
        \item According to the NeurIPS Code of Ethics, workers involved in data collection, curation, or other labor should be paid at least the minimum wage in the country of the data collector. 
    \end{itemize}

\item {\bf Institutional review board (IRB) approvals or equivalent for research with human subjects}
    \item[] Question: Does the paper describe potential risks incurred by study participants, whether such risks were disclosed to the subjects, and whether Institutional Review Board (IRB) approvals (or an equivalent approval/review based on the requirements of your country or institution) were obtained?
    \item[] Answer: \answerNA{} 
    \item[] Justification: Evaluating pathologists were members of the research team.
    \item[] Guidelines:
    \begin{itemize}
        \item The answer \answerNA{} means that the paper does not involve crowdsourcing nor research with human subjects.
        \item Depending on the country in which research is conducted, IRB approval (or equivalent) may be required for any human subjects research. If you obtained IRB approval, you should clearly state this in the paper. 
        \item We recognize that the procedures for this may vary significantly between institutions and locations, and we expect authors to adhere to the NeurIPS Code of Ethics and the guidelines for their institution. 
        \item For initial submissions, do not include any information that would break anonymity (if applicable), such as the institution conducting the review.
    \end{itemize}

\item {\bf Declaration of LLM usage}
    \item[] Question: Does the paper describe the usage of LLMs if it is an important, original, or non-standard component of the core methods in this research? Note that if the LLM is used only for writing, editing, or formatting purposes and does \emph{not} impact the core methodology, scientific rigor, or originality of the research, declaration is not required.
    \item[] Answer: \answerYes{} 
    \item[] Justification: We use LLMs as part of our data pipeline, described in Appendix Section~\ref{llm_usage}.
    \item[] Guidelines:
    \begin{itemize}
        \item The answer \answerNA{} means that the core method development in this research does not involve LLMs as any important, original, or non-standard components.
        \item Please refer to our LLM policy in the NeurIPS handbook for what should or should not be described.
    \end{itemize}

\end{enumerate}

\end{document}